\documentclass{article}

\usepackage[english]{babel}

\usepackage[letterpaper,top=2cm,bottom=2cm,left=3cm,right=3cm,marginparwidth=1.75cm]{geometry}

\usepackage[section]{placeins}
\usepackage[colorlinks=true, allcolors=blue]{hyperref}
\usepackage{svg}

\usepackage{amsmath,bm,graphicx,psfrag,epsf,enumerate,placeins,float,sectsty,amstext,relsize,wrapfig,listings,subfigure,amsfonts,amsthm,mathabx,enumitem,hyperref,grffile,colortbl,dsfont,bbm,setspace, float, xcolor, multirow}
\usepackage{cleveref}
\usepackage[compress]{natbib}

\renewcommand\cite{\citep}

\newcommand{\beginsupplement}{ 
        \setcounter{section}{0}
        \renewcommand{\thesection}{S\arabic{section}} %
         \renewcommand{\thesubsection}{\thesection.\arabic{subsection}}
        \setcounter{table}{0}
        \renewcommand{\thetable}{S\arabic{table}} %
        \setcounter{figure}{0}
        \renewcommand{\thefigure}{S\arabic{figure}} %
     }

\newcommand{\score}{\ensuremath{s}}
\newcommand{\scoreEg}[1]{\ensuremath{\score_{#1}^*}}
\newcommand{\scoreEgCl}[2]{\ensuremath{\score_{#1}^{#2}}}
\newcommand{\scoreVec}[1]{\ensuremath{\mathbf{\score}_{#1}}}

\allowdisplaybreaks  

\title{Identifying Incorrect Annotations  \\ in Multi-Label Classification Data}
\author{Aditya Thyagarajan \hspace*{17mm} Elías Snorrason \\
\hspace*{2mm} \texttt{aditya@cleanlab.ai} \hspace*{10mm}
\texttt{elias@cleanlab.ai} 
\\[1em]
Curtis Northcutt \hspace*{22mm}  Jonas Mueller \\ \hspace*{2mm} \texttt{curtis@cleanlab.ai} \hspace*{12mm} \texttt{jonas@cleanlab.ai}
\\[1em] 
\emph{Cleanlab}}
\date{} 

\begin{document}
\maketitle

\begin{abstract}
In multi-label classification, each example in a dataset may be annotated as belonging to one or more classes (or none of the classes). Example applications include image (or document) tagging where each possible tag either applies to a particular image (or document) or not. With many possible classes to consider, data annotators are likely to make errors when labeling such data in practice. 
Here we consider algorithms for finding mislabeled examples in multi-label classification datasets. We propose an extension of the Confident Learning framework to this setting, as well as a label quality score that ranks examples with label errors much higher than those which are correctly labeled. Both approaches can utilize \emph{any} trained classifier. 
After demonstrating that our methodology\footnote{Code to run our method: \url{https://github.com/cleanlab/cleanlab} \\ \hspace*{4.5mm} Code to reproduce our benchmarks: \url{https://github.com/cleanlab/multilabel-error-detection-benchmarks}}
empirically outperforms other algorithms  for label error detection, we apply our approach to discover many label errors in the CelebA image tagging dataset. 
\end{abstract}

\section{Introduction}

Many real-world datasets contain label errors, which should be identified and fixed to train the best models for supervised learning  \citep{natarajan2013learning, lee2017cleannet, song2022survey, huang2019o2u, northcutt2021labelerrors}. 
Label errors are particularly likely in tasks where annotators must make many choices for each individual example in the dataset, such as structured prediction \citep{reiss2020identifying}. 
While fixing these errors must be done carefully to avoid introducing bias in the dataset, finding the mislabeled examples can be done much more efficiently with the help of algorithms \citep{kuan2022}. 
Confident Learning algorithms offer a principled approach to detect mislabeled examples in multi-class classification datasets \citep{northcutt2021confidentlearning}. Alternatively one may rank the examples in a dataset by their likelihood of being mislabeled, as estimated via a \emph{label quality score} \citep{kuan2022}, which enables efficient review of the most questionable labels.  

\begin{figure}[!b]
    \centering
    \includegraphics[width=\textwidth]{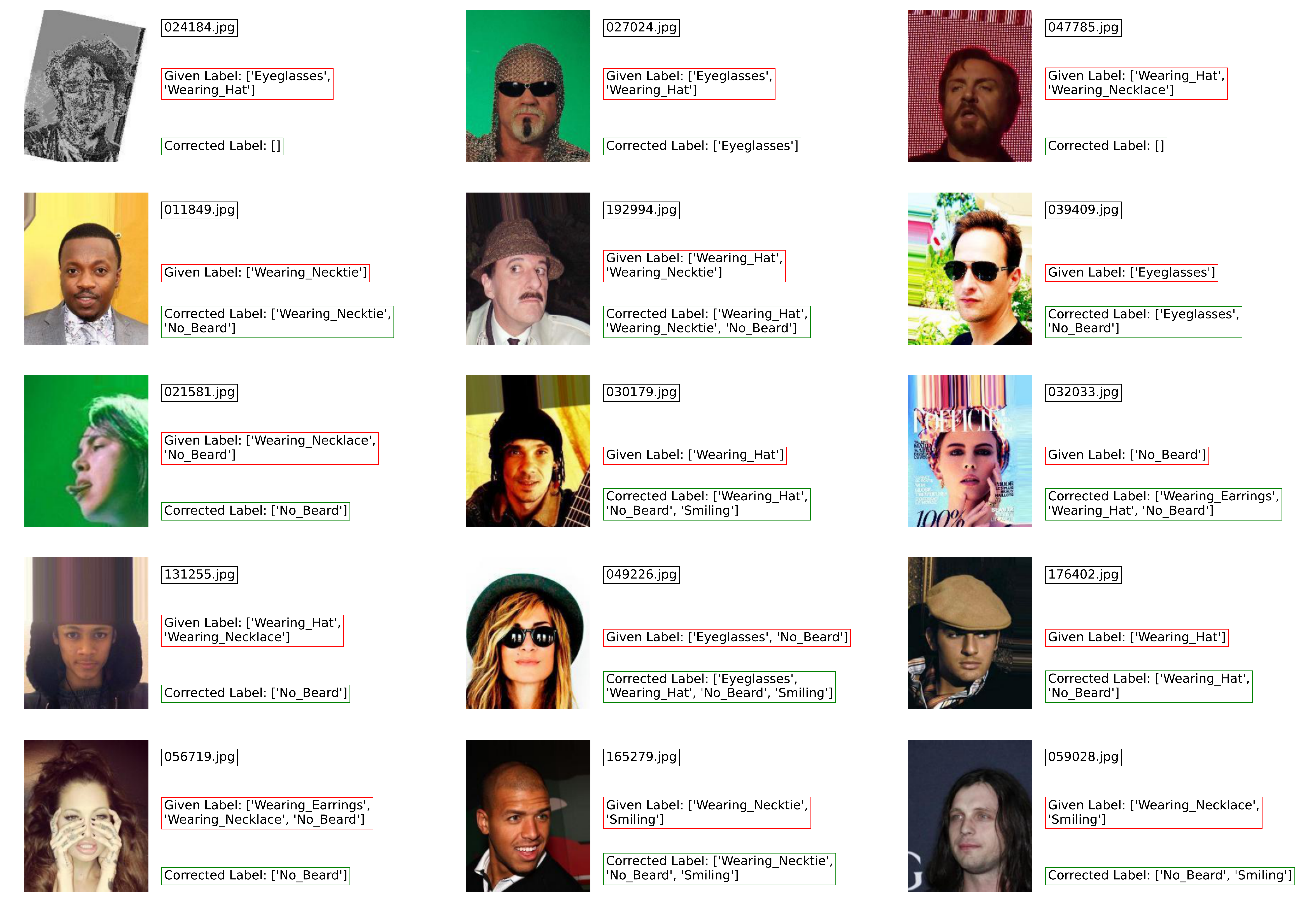}
    \vspace*{-0.5cm}
    \caption{15 of the top-50 scoring images in the CelebA dataset based our EMA label-quality-score, which have also been flagged by our Confident Learning extension. Here we choose not show many additional flagged label errors related to incorrect annotation of the \texttt{no\_beard} tag amongst these top-50 images, in order to highlight diverse label errors in this dataset automatically detected by our methodology. There are both extraneously added tags as well as many missing tags in the dataset.
    }
    \label{fig:celeba}
\end{figure}

While detection of label errors has been studied for \emph{multi-class} classification tasks \citep{brodley1999identifying,muller2019mislabeled,northcutt2021confidentlearning}, and for other tasks like token classification \citep{token, klie2022annotation} or image  segmentation \citep{segmentation}, there has been little research on the task of \textbf{label error detection} in \emph{multi-label} classification datasets. 
In multi-class classification, each example belongs to exactly one of $K$ classes. This mutually exclusive restriction is lifted in multi-label classification, where each example can belong to one or more of the $K$ classes, or none of the  classes at all. This task can be equivalently framed in a \emph{one-vs-rest} fashion as a series of $K$ binary classification problems, where each class either applies to a particular example or not.
Although the one-vs-rest framework serves as a useful way to think about multi-label classification, it is important to model correlations between the classes to produce the best possible predictions. While the term \emph{one-vs-rest} is commonly used to refer to a strategy of training a separate classifier independently for each binary classification problem, we do not advocate for such an approach that ignores inter-class correlations, and write  \emph{one-vs-rest} to refer to a general viewpoint of multi-label classification where each class either applies or not to each example in the dataset. For instance in an image tagging application, whether or not a person is wearing goggles is likely correlated with whether or not they are wearing a beanie. The given labels in Figure \ref{fig:celeba} provide an example of real multi-label classification data.

From the one-vs-rest perspective, it is clear that annotators for multi-label classification data must consider $K$ different questions when labeling each example. Thus the potential for annotation errors is high, particularly when examples are rapidly annotated and $K$ is large. 
Here we consider label error detection for multi-label classification datasets via two approaches. The first approach estimates \emph{which} examples are mislabeled, and the second approach estimates a label quality score for each example that should be monotonically related to the likelihood it is correctly labeled. \citet{klie2022annotation} call these two approaches \emph{Flagger} and \emph{Scorer} methods, with the former being useful to estimate the number of mislabeled examples in the dataset and other statistics of the label errors (such as which class is often erroneously substituted in place of another class), and the latter being useful to rank the examples and  prioritize which ones should be re-examined under a limited  label verification budget.

Our algorithms for label error detection are entirely model-agnostic and can utilize any multi-label classifier, regardless how it was trained. 
In summary, the contributions of this paper include:
\begin{enumerate}[wide, labelwidth=!, labelindent=0pt]
    \item A generalization of Confident Learning to multi-label classification tasks, 
    enabling estimation of (and learning with) label issues in multi-label datasets.
    \item A novel quality score for each example's annotated label in multi-label classification datasets. 
    \item Characterizing incorrectly tagged images in the  CelebA dataset (we estimate there are over $30,000$). 
\end{enumerate}

\section{Preliminaries}

Here we assume a multi-label classifier has already been trained and we have obtained  predicted class probabilities $p_i \in \mathbb{R}^K$ for each example $i$ in our dataset. 
To remain robust against overfitting, these predictions should be out-of-sample, meaning $p_i$ is produced by a copy of the model that has never been trained on the $i$th example, which is easily achieved using cross-validation.
Note that the $K$ probability values in any  $p_i$ need not sum to 1, since the classes are not mutually exclusive in multi-label classification.

A major strength of our proposed methods is how straightforward they are. The sole inputs required are these predicted class probabilities and the given labels for the dataset. Access to the original feature values is not required, nor is any form of nonstandard modeling. Given more accurate predicted class probabilities, our label error detection methods can immediately better identify the annotation errors in a dataset.

\section{Extending Confident Learning to Flag Label Issues in \\ Multi-labeled Data} 
\label{sec:flag}

We first consider estimating exactly which examples in the dataset are mislabeled. 
For multi-class classification, \citet{northcutt2021confidentlearning} proposed \emph{Confident Learning}, an approach theoretically proven to identify the mislabeled examples under mild assumptions, even in the presence of asymmetric class-conditional label noise and an imperfectly trained classifier. 
Despite only requiring predicted class probabilities (from any trained classifier) to identify label errors, 
Confident Learning has been shown to empirically outperform many more complex approaches  \cite{northcutt2021confidentlearning, klie2022annotation}. 
Thus, we consider an extension of this framework to multi-label datasets. 
A naive extension could be to consider each observed combination of classes that co-occur for a particular example as a separate ``class'', which would then be mutually exclusive of the others as in multi-class classification (such that Confident Learning can be applied). However this can result in up to $2^K$ ``classes'' and is statistically/computationally impractical.

Instead here we adopt the one-vs-rest perspective. For each of the $K$ classes, we  form a separate binary label $b_{i}^{k}$ whether class $k$ applies to example $i$ or not, according to the given annotation. We extract the corresponding $k$th entry from the predicted class probabilities $p_i$ (denoted as $p_{i}^{k}$) as an estimate of the probability that $b_{i}^{k} = 1$. 
For each class $k$, Confident Learning can be independently applied to these binary labels and predicted binary probabilities $p_{i}^{k}$ to flag in which examples the annotated binary label $b_{i}^{k}$ is estimated to be wrong. Finally, we can simply report the \emph{union} of these flagged examples over all classes $k=1,...,K$ as the subset of examples in the multi-label classification dataset whose given label contains an error.

This approach entails a straightforward extension of Confident Learning to multi-label settings. All of the theoretical guarantees established by \citet{northcutt2021confidentlearning} hold for our extension as well, as long as the number of classes $K$ is significantly lower than the number of examples in the dataset $N$ (which is almost always the case to train a reasonable classifier). 
Note that this approach does not require additional complex modeling of dependencies between classes. We assume the multi-label classifier which has been trained to produce the $p_i$ already reflects these dependencies in its predictions (i.e.\ we caution \emph{against} training the classifier in a one-vs-rest manner). Even if these dependencies are imperfectly captured in $p_i$, this approach should be able to accurately flag the examples whose labels are not entirely correct as long as the per-class binary probabilities $p_{i}^{k}$ are good enough estimates of the binary label $b_{i}^{k}$ that satisfy the conditions of \citet{northcutt2021confidentlearning}. If multiple classes tend to all be mis-annotated in the same examples, our approach should be able to  correctly flag these examples, even though the modeling of dependencies between classes is left as the responsibility of the multi-label classifier. As for Confident Learning, the efficacy of our approach does depend on the accuracy of the trained classifier, and in multi-label settings, a classifier that effectively models dependencies between classes will usually produce more accurate predictions.

\section{Methods for Scoring Label Quality in Multi-Labeled Data}
\label{sec:score}

A limited budget may prohibit verification of the labels of all candidate examples flagged by our Confident Learning extension. For such settings, it is important to prioritize which examples should be looked at more closely. Here we consider numeric \emph{label quality scores} that can be used to rank the examples by our estimated confidence that each one is correctly labeled \citep{kuan2022}. Under the one-vs-rest perspective, each example can be considered to have $K$ binary labels $b_{i}^{k}$, and in principle one could compute a label quality score $s_{i}^{k}$ for each of these $b_{i}^{k}$ to find which individual $b_{i}^{k}$ are incorrect. However verifying whether $b_{i}^{k}$ is correct or not for a particular example $i$ and class $k$ requires investing time to understand the semantics of example $i$. It is thus more efficient to simultaneously verify the other binary labels for this same example $i$ than to review a different example. 

Hence a key focus of this paper is to estimate \textbf{a single label quality score $\scoreEg{i}$ for each multi-label example in a dataset}, where $\scoreEg{i}$ should take smaller values for those examples where \emph{any} of the $b_{i}^{k}$ are incorrect (i.e.\ the annotated label is not entirely right).
Our primary aim is to ensure these label quality scores achieve high precision/recall for detecting those examples that have any annotation error. 
A secondary aim is to ensure the lowest $\scoreEg{i}$ values in the dataset correspond to severely mislabeled examples for which multiple $b_{i}^{k_1}, b_{i}^{k_2}, \dots$ are incorrect. 
Using such severely mislabeled examples to train a model can have more damaging effects than examples $i$ for which only one $b_{i}^{k}$ is incorrect (see Table \ref{tab: model accuracies}), and thus it is more critical to spot these cases.

Favoring a straightforward approach, we propose to compute $\scoreEg{i}$ by first computing a separate label quality score $s_i^k$ for each binary label $b_i^k$ corresponding to a particular class $k$, and subsequently pooling $s_i^1, ..., s_i^K$ into a single overall score for example $i$. 
For a particular class $k$, each $b_i^k$ corresponds to a standard binary classification and thus any label quality score for multi-class classification can be employed to compute $s_i^k$. Here we mainly focus on one effective score studied by \citet{kuan2022,  northcutt2021confidentlearning, klie2022annotation} called \emph{self-confidence}, which is simply the classifier-estimated likelihood of the given label: 
\begin{equation}
    s_i^k = b_i^k \cdot p_i^k + (1-b_i^k) \cdot (1-p_i^k)
\end{equation}

\subsection{Methods to Pool Scores}

We have several methods to pool the $K$ scores $\scoreVec{i} \equiv (\scoreEgCl{i}{1}, ... , \scoreEgCl{i}{K})$ into a single score, $\scoreEg{i}$, for the $i$th example in a dataset. Recall we would like $\scoreEg{i}$ to have high precision/recall for detecting those examples $i$ for which \emph{any} of the binary labels $b_i^1,..., b_i^K$ are incorrect, and we also prefer $\scoreEg{i}$ to take the lowest values if many of the binary labels $b_i^1,..., b_i^K$ are incorrect.
The following four baseline pooling methods simply use the lowest, highest, mean, or median score as the aggregated value.
\begin{equation}
\scoreEg{i} = \min_k  \{ \scoreEgCl{i}{k} \}
\label{eq:minpool}
\end{equation}
\begin{equation}
\scoreEg{i} = \max_k \{ \scoreEgCl{i}{k} \}
\end{equation}
\begin{equation}
\scoreEg{i} = \mathop{\text{mean}}_k \{ \scoreEgCl{i}{k} \}  
\end{equation}
\begin{equation}
\scoreEg{i} = \mathop{\text{median}}_k \{ \scoreEgCl{i}{k} \}
\end{equation}
Intuitively, the min-pooled score can be viewed as a bound on the probability that at least one of $b_i^1, ..., b_i^K$ is incorrect. Thus this baseline should achieve better precision/recall than these other baselines, but will not necessarily prioritize examples for which many of $b_i^1, ..., b_i^K$ are incorrect. In contrast, mean/median pooling lead to scores that better reflect all of $\scoreEgCl{i}{1}, ..., \scoreEgCl{i}{K}$, which can be helpful to prioritize examples with many errors in $b_i^1, ..., b_i^K$, but the resulting $\scoreEg{i}$ are also more sensitive to nuisance variation (e.g.\ due to estimation error) in the $s_i^1, ..., s_i^K$ for correctly annotated $b_i^1, ..., b_i^K$.

\subsection{Exponential Moving Average (EMA)}

To achieve a score that  accounts for all classes' scores but emphasizes the score of the most suspect class annotation, we propose to pool $s_i^1,...,s_i^K$ via an exponential moving average of these scores, after they have been sorted in descending order. 
Letting $\check{s}_i^1 \ge \dots \ge \check{s}_i^K$ denote the same scores $s_i^1,...,s_i^K$ now sorted in decreasing order (separately for each example $i$), we run through them in this order and report their exponential moving average as an overall label quality score:
\begin{equation}
\scoreEg{i} = S^K_i, \text{ where }  
S^t_i = 
  \begin{cases} 
   \check{s}_i^1 \quad & \text{ for } t = 1 \\
   \alpha \cdot \check{s}_i^{t} + (1 - \alpha) \cdot S^{t-1}_i \quad & \text{ for } t > 1
  \end{cases}
\end{equation}
Here $0 < \alpha < 1$ is a forgetting-factor constant that is fixed a priori to define the relative importance of the $s_i^k$ based on where they occur in the sorted ordering. 
We propose to use $\alpha = 0.8$, such that $\scoreEg{i}$ is most heavily influenced by the smallest $s_i^k$ (corresponding to the classes $k$ whose binary label $b_i^k$ is most likely incorrect), but also still directly affected by all $s_i^1, ..., s_i^K$ values. This aligns with our primary aim to ensure $\scoreEg{i}$ achieves the highest precision/recall for detecting examples with any error in $b_i^1,..., b_i^K$ while also favoring examples for which many of these binary labels are incorrect. Figures \ref{fig:alpha1}-\ref{fig:alpha4} show the effect of different $\alpha$ values.

\subsection{Additional Pooling Methods}
Many alternative pooling methods could be applied to obtain a label quality score per example. Here we consider additional approaches evaluated in our benchmark study presented in Section \ref{sec:benchmarkscores}.

\paragraph*{Softmin pooling.}
Recall that pooling via a hard minimum as in (\ref{eq:minpool}) implies $\scoreEg{i}$ is solely determined by one of the $s_i^k$ values for each example. An alternative way to take the other classes' scores into consideration while still emphasizing the minimum, as done by our EMA method, is to pool via a softer minimum-like operator. Recalling that $\mathbf{s}_i = (s_i^1, ..., s_i^K)$ denotes a vector of the per-class scores for example $i$, a simple way to pool using a softer minimum is via the softmax operator as follows:
\begin{equation}
\scoreEg{i} = \scoreVec{i} \cdot \text{softmax}_\tau ( 1 - \scoreVec{i} ) = \frac{ \sum_{k=1}^K s_i^k \cdot \exp \big( \frac{1-\scoreEgCl{i}{k}}{\tau} \big)}{\sum_{k=1}^K \exp \big(\frac{1-\scoreEgCl{i}{k}}{\tau} \big)}
\end{equation}
Here $\tau > 0$ denotes the \emph{temperature} of the softmax. We choose $\tau = 0.1$ which heavily emphasizes the smallest scores corresponding to the most suspicious class annotations. Larger values of $\tau$ did not perform as well in our benchmarks.  

\paragraph*{Log-transform Pooling.}
Alternatively, taking an arithmetic mean of the logarithm of all the scores should significantly emphasize the low-scoring class for a particular example, while still being accounting for the remaining classes' scores. 
\begin{equation}
\scoreEg{i} = \frac{1}{K} \sum_{k=1}^K  \log(\scoreEgCl{i}{k}  + \epsilon) 
\end{equation}
Here a positive infinitesimal, $\epsilon = 1e-8$, is added to all scores to avoid scores of zero. 
This $\scoreEg{i}$ corresponds to a common loss function used to train multi-label classifier models (log-likelihood). Much existing research has proposed using the training loss of each example as a label quality score  \cite{klie2022annotation, muller2019mislabeled}, as done via this log-transform approach.

\paragraph*{Cumulative Average of Bottom Scores.}
For subsequent pooling methods, we let $\hat{s}_i^1 \le ... \le \hat{s}_i^K,$ denote the same values as $s_i^1, ..., s_i^K$ now sorted in increasing order. 
Another pooling method that emphasizes the smallest scores, but not only the minimum value, is to take an average of the $J$ smallest scores for each example. Here we simply set $J = 2$ as larger values did not perform as well. 
\begin{equation}
\scoreEg{i} = \frac{1}{J} \sum_{k=1}^J \hat{s}_{i}^k 
\end{equation}

\paragraph*{Weighted Sum of Cumulative Averages.}
A more flexible variant of this method is to take cumulative averages across the per-class scores (again sorted in ascending order) for different values of $J$, and subsequently report an exponentially-weighted sum of these averages as the aggregated score.
\begin{equation}
    \scoreEg{i} = \sum_{J=1}^K \sum_{k=1}^J \frac{\exp(1-J)}{J} \hat{s}_i^k
\end{equation}


\paragraph*{Mean Simple Moving Average (SMA).}
Most of the presented pooling methods aim to emphasize lower per-class scores based on the intuition that these should matter more to detect examples $i$ for which any of $b_i^1,..., b_i^K$ are misannotated. De-emphasizing the other classes' scores helps mitigate nuisance variation in the scores for class annotations that are likely correct. 
Another way to mitigate nuisance variation may be to smooth the scores across different classes. 
We consider simple moving averages of the sorted scores with period $P < K$. Moving averages smooth out variation in the scores for adjacent classes (in the sort ordering), which are often both correctly annotated. 
These moving averages can then be mean-pooled as an overall label quality score.
\begin{equation}
\scoreEg{i} = \frac{1}{P (K-P+1)} \sum_{k'=P}^{K}  \sum_{k=k'-P+1}^{k'} \hat{s}_{i}^{k} 
\end{equation}
Empirically, we found $P=2$ outperformed larger values of $P$ in our benchmarks.

\section{Experiments}

\subsection{Evaluating label quality scores}
\label{sec:benchmarkscores}
For evaluation, we consider two groups of multi-label classification datasets (one \emph{large}, one \emph{small}) that contain examples with multiple noisy labels and bag of words features (these datasets are detailed in Appendix \ref{sec:datadetails}). Each group has ten different datasets, generated from the same underlying distribution. Between the two groups, datasets differ in their overall sample size,
feature- and class counts, the expected number of classes  which apply to each example (i.e.\ fraction of $b_i^k$ which equal 1), and overall degree of label noise. We know the ground truth labels for all datasets and can thus evaluate the performance of different label error detection methods.

To produce predicted class probabilities for finding label errors, we train two multi-label classifiers, a logistic regression model and a random forest model, on each dataset. Both classifiers are trained using one-vs-rest 5-fold cross-validation to produce held-out predictions for every example in each dataset. 
The random forest model tends to be more accurate than the logistic regression model for the \emph{Small} datasets, but its predicted class probabilities vary less smoothly across different feature values (Table \ref{tab: model accuracies}). The differences in between these models' accuracies for the \emph{Large} datasets is much lower, where classes are more likely to be linearly separable.


By comparing ground truth labels with noisy given labels for each dataset, we evaluate label quality scores for each example (estimated from the given labels) via various precision/recall metrics. For each metric listed below, higher values correspond to a better performing method. 

\paragraph{AUPRC.} 
The area under 
Precision-Recall 
curve obtained by comparing estimated scores $s^*_i$ against a binary target whether the given label for example $i$ contains an annotation error or not.  

 \paragraph{AP @ T.} The Average Precision of the label quality score $\scoreEg{i}$ in detecting mislabeled examples among the bottom-$1$, bottom-$2$, ..., bottom-$T$  scoring examples in the dataset. Note that $\textbf{AUPRC} = \textbf{AP @ }N$ where $N$ is the number of examples in the dataset. 
    We report \textbf{AP @ T} 
    for $T$ equal to the number of actual mislabeled examples in each dataset. This threshold based on the number of relevant items is a popular  way to evaluate information retrieval systems.

\paragraph{Spearman Correlation.} 
The aforementioned evaluation metrics measure how well our scores can retrieve examples for which \emph{any} of the per-class binary labels $b_i^1,...,b_i^K$ is incorrect. To measure how well our scores retrieve severely mislabeled examples for which \emph{many} of these binary labels are incorrect, we report the Spearman's rank correlation between $s_i^*$ and the number of $b_i^1,...,b_i^K$ which are incorrect over all examples $i$ in a dataset.

\paragraph{Average 2-Precision @ T.} Another way to evaluate the identification of severely mislabeled examples is by only considering positive hits to be those examples $i$ for which at least 2 of the per-class annotations $b_i^1,...,b_i^K$ are incorrect. We can subsequently report the Average Precision @ T for our how well our scores prioritize such positives before other examples. We also report {\textbf{Average 3-Precision @ T}} where positive hits are defined only as those examples for which at least 3 of the per-class annotations $b_i^1,...,b_i^K$ are incorrect.

\begin{figure}[tb] 
\begin{center}
\textbf{Logistic Regression} \\[0.1cm]
\includegraphics[width=0.7\textwidth, trim={5cm 5cm 5cm 5cm},clip]{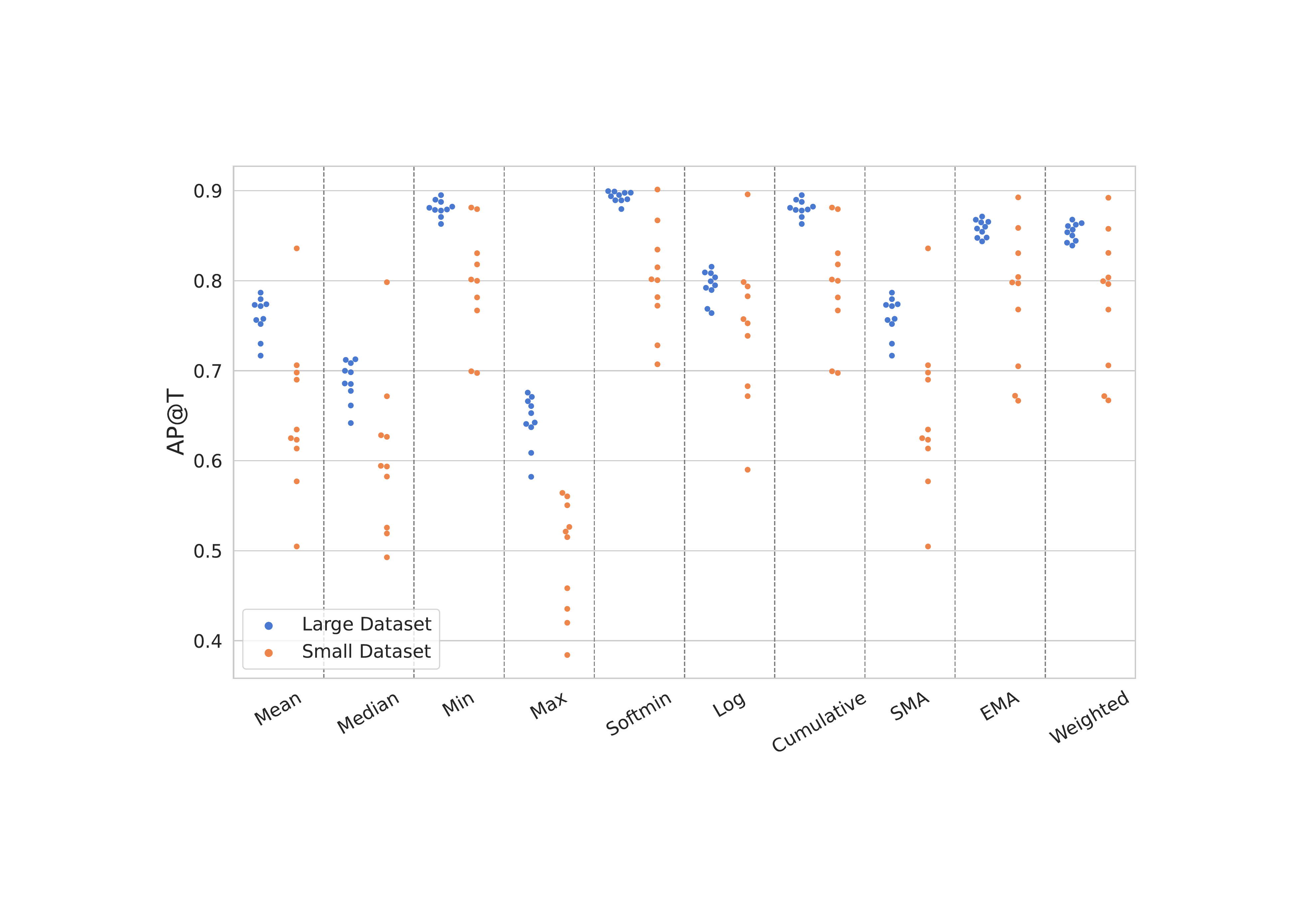}
\\[0.1cm]
\textbf{Random Forest} \\[0.1cm]
\includegraphics[width=0.7\textwidth, trim={5cm 5cm 5cm 5cm},clip]{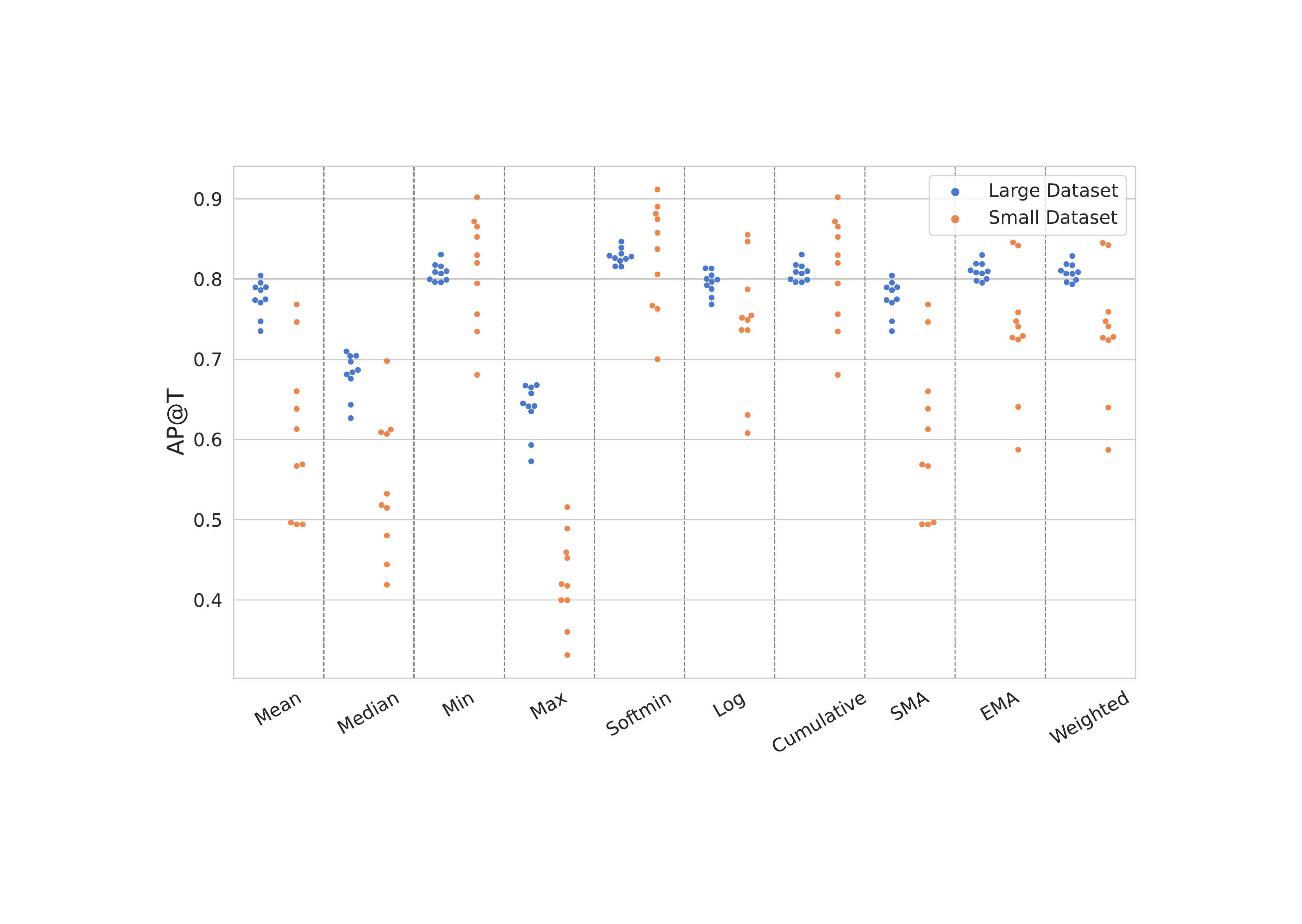}
\end{center}
\vspace*{-0.6cm}
\caption{Average Precision @ $T$ achieved by various overall label quality scores $\scoreEg{i}$ for each dataset (shown as dot), where $T$ is the number of mislabeled examples in the dataset.
We show results based on predicted class probabilities from both a Logistic Regression model and a Random Forest model.}
\label{fig:ap_numerr}
\vspace*{2em}
\end{figure}

\begin{figure}[tb] 
\begin{center}
\textbf{Logistic Regression} \\[0.1cm]
\includegraphics[width=0.7\textwidth, trim={5cm 5cm 5cm 5cm},clip]{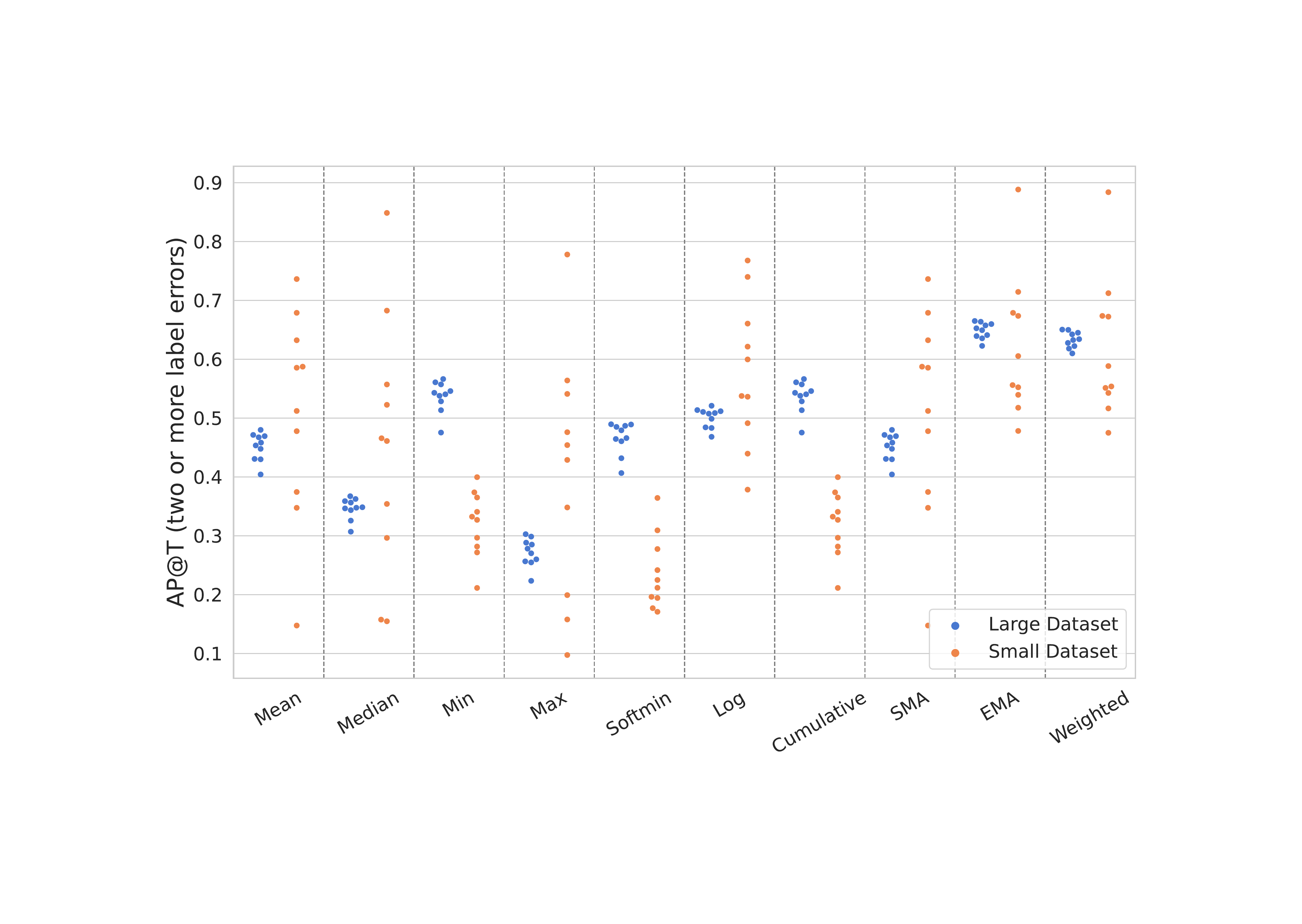}
\\[0.1cm]
\textbf{Random Forest} \\[0.1cm]
\includegraphics[width=0.7\textwidth, trim={5cm 5cm 5cm 5cm},clip]{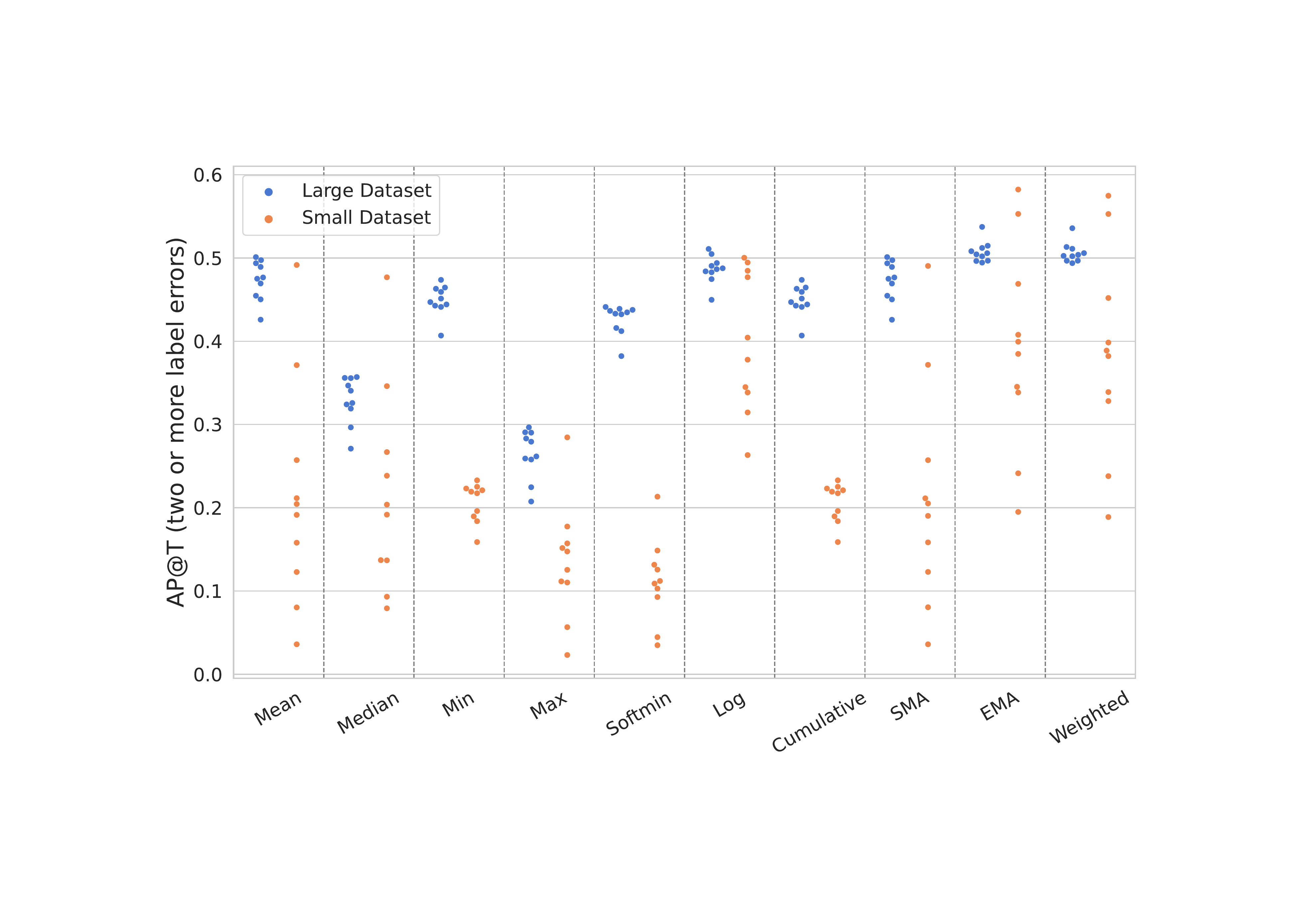}
\end{center}
\vspace*{-0.6cm}
\caption{Average 2-Precision @ $T$ achieved by various overall label quality scores $\scoreEg{i}$ for each dataset, where $T$ is the number of mislabeled examples in the dataset.
We show results based on predicted class probabilities from both a Logistic Regression model and a Random Forest model.}
\label{fig:ap2_numerr}
\end{figure}

\paragraph{Results.} 
\Cref{fig:ap_numerr,fig:ap2_numerr,fig:auprc,fig:ap3_numerr,fig:spearman} show that across all metrics, EMA performs well relative to the other overall label quality scoring approaches. No other approach consistently performs well across the different metrics, particularly because some approaches like min/softmin-pooling are good at detecting mislabeled examples but not those which are severely mislabeled, whereas the situation is reversed for other methods like cumulative average pooling. As expected intuitively, weighted cumulative average pooling performs similar to EMA, but is not quite as effective overall.

These results validate that one can better detect examples with any sort of label error via methods like min-pooling which are not influenced by most of the per-class label quality scores for a particular example. However accounting for more per-class scores beyond just the minimum score helps better detect severely mislabeled with many errors in more than one of their per-class annotations. Of the alternative methods, those most similar to EMA (like our weighted cumulative average and softmin pooling) perform the best. Using a typical multi-label classification training loss (equivalent to log transform pooling) is not as effective for overall label quality scoring as EMA and similar approaches.

\Cref{fig:alpha1,fig:alpha2,fig:alpha3,fig:alpha4} reveal the effect of different $\alpha$ values in our EMA label quality scoring method.
As $\alpha$ grows toward 1, the overall EMA score $\scoreEg{i}$ becomes increasingly dominated by the smallest of the per-class scores $s_i^1, ..., s_i^K$. Our benchmarks reveal this generally leads to better detection of examples where \emph{any} of the per-class annotations $b_i^1,..., b_i^K$ are incorrect (Figure \ref{fig:alpha1}) but worse detection of the severely mislabeled examples for which many of $b_i^1,..., b_i^K$ are incorrect (Figures \ref{fig:alpha2}-\ref{fig:alpha4}). $\alpha = 0.8$ appears to effectively address  both objectives. Since the weight of the 
$k$-th smallest per-class score $\hat{s}_i^k$ is $\alpha(1 - \alpha)^{k-1}$ in the final moving average, with $\alpha = 0.8$, the overall EMA score $\scoreEg{i}$ is 3.2\% determined by the 3rd lowest per-class score $\hat{s}_i^3$ and only 0.64\% determined by the 4th lowest per-class score $\hat{s}_i^4$.


\subsection{Finding label errors in the CelebA image tagging dataset}
To demonstrate our approach in an image tagging application, we consider a subset of the popular CelebA dataset \citep{liu2015faceattributes}.
CelebA is a face attributes dataset depicting images of celebrities labeled with various attributes. Here we consider only the following subset of the original CelebA tags (and the subset of 188,000 images annotated with at least one of the tags under consideration): \texttt{Wearing\_Hat}, \texttt{Wearing\_Necklace}, \texttt{Wearing\_Necktie}, \texttt{Eyeglasses}, \texttt{No\_Beard}, and \texttt{Smiling}.

We train a neural network for multi-label classification by fine-tuning a pretrained network backbone (efficientnet \citep{tan2019efficientnet} implemented in the TIMM library \citep{rw2019timm}) with a $K$-dimensional linear output layer added for our prediction task. Each output note uses an independent sigmoid activation rather than a softmax activation which would otherwise constrain the predicted class probabilities to be mutually exclusive. 
Rather than training this classifier in a one-vs-rest fashion, we fine-tune the model using Adam \citep{kingma2015adam} to jointly optimize a binary cross-entropy loss at the output note for each class, such that the classifier can learn to model correlations between tags.  We produce held-out predictions for each image in the dataset via 4-fold cross-validation. 

\begin{figure}[tb] 
\begin{center}
{\large Inconsistency in  \texttt{Wearing\_Hat} Annotations} \\
\includegraphics[width=0.95\textwidth]{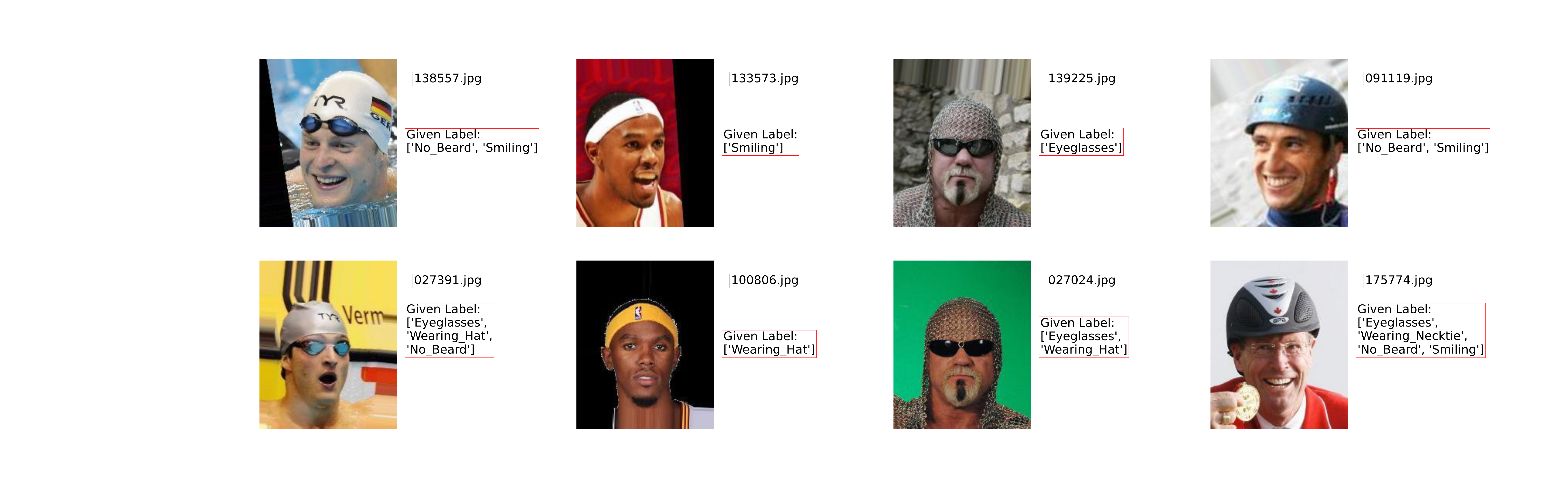}
\\[0.3cm]
{\large
Inconsistency in  \texttt{No\_Beard} Annotations} 
\\[0.1cm]
\includegraphics[width=0.99\textwidth]{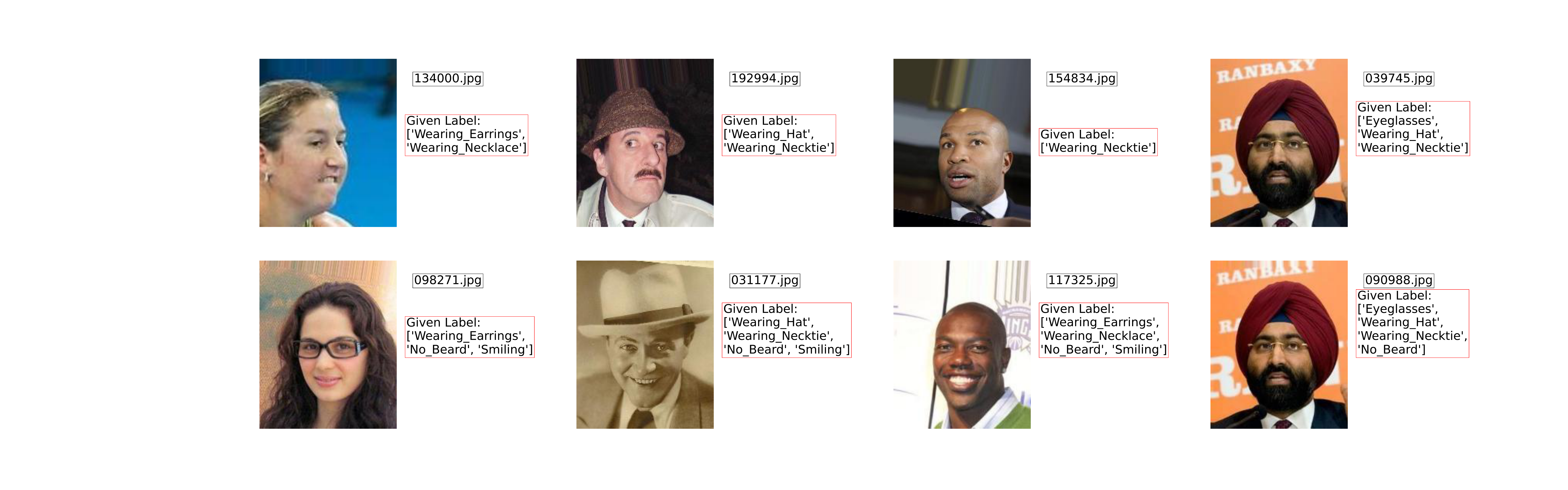}
\\[0.3cm]
{\large 
Inconsistency in 
\texttt{Eyeglasses} Annotations
}
\\ 
\includegraphics[width=0.68\textwidth]{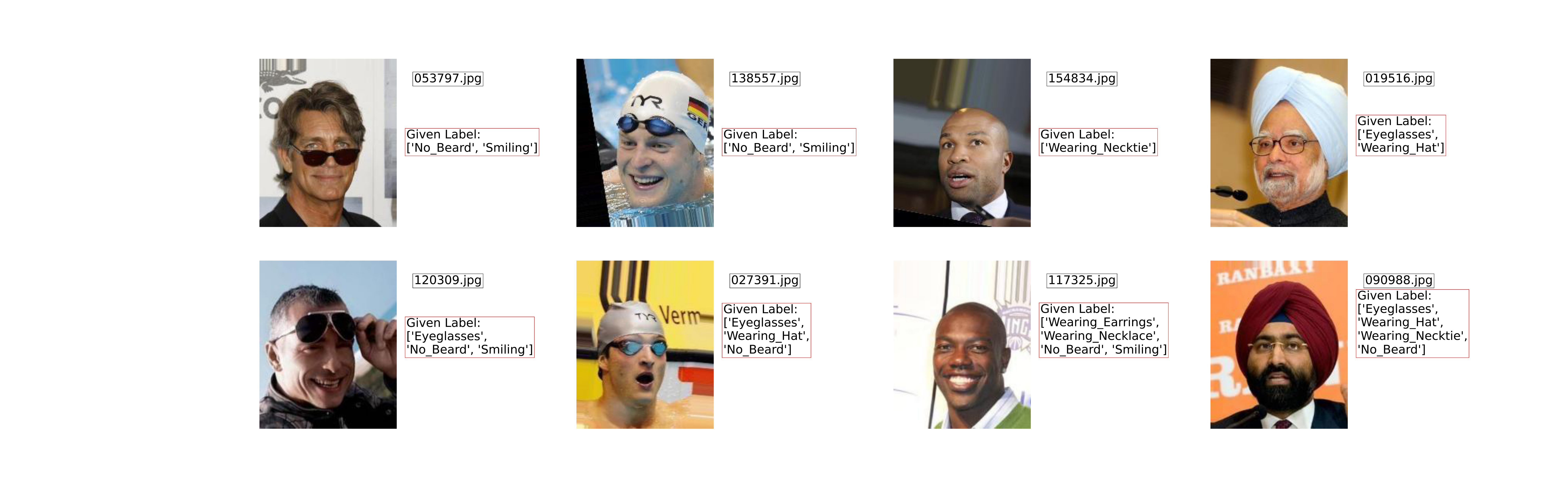}
\end{center}
\vspace*{-0.5cm}
\caption{Examples of inconsistency in  \texttt{Wearing\_Hat}, \texttt{No\_Beard}, 
or 
\texttt{Eyeglasses} 
annotations in the CelebA dataset. Images shown are selected among the smallest (most likely erroneous) EMA label quality scores. These examples represent a small subset of the many issues in these tags that we observed throughout the CelebA dataset.
}
\label{fig:inconsistencies}
\end{figure}

Running our extension of Confident Learning and sorting the flagged examples based on our EMA label quality score reveals that CelebA contains many mislabeled examples (Figure  \ref{fig:celeba}). There are both extraneously added tags as well as many missing tags in the dataset. In particular, CelebA contains a \texttt{No\_Beard} tag that should actually apply to a large fraction of the images in the dataset, yet is often not present in the given label. The annotations contain may other inconsistencies reflecting likely annotator confusion regarding when tags like \texttt{Wearing\_Hat} or \texttt{Eyeglasses} should apply (Figure \ref{fig:inconsistencies}).

We manually review 100 randomly chosen images from the dataset very carefully and estimate 15 of them to be mislabeled. This 15\% estimate implies there are around 30,000 mislabeled images in the full CelebA dataset!
Among the top-100 images ranked according to our EMA label quality score $\scoreEg{i}$, we identify 67 mislabeled images. Thus the Lift @ 100 for detecting annotation errors via our approach is around 4.5, i.e.\ \textbf{mislabeled images are 4.5 times more prevalent among the set prioritized by our label quality score compared to the overall dataset}. An effective label quality score can help us identify the annotation errors in a dataset far more efficiently than random inspection.

\section{Discussion}

This paper introduced model-agnostic methods to identify which examples are mislabeled in multi-label classification dataset and score the confidence of these estimates. A key question to extend these capabilities from multi-class classification to multi-label settings is how to define a single overall label quality score per example. Our proposed EMA score effectively detects  examples whose annotated labels contain errors,  prioritizing severely mislabeled examples whose label contains many errors. 

Beyond images, the approach presented here can be used to easily find such annotation errors in any type of multi-label classification dataset, as long as a reasonable classifier can be fit to the data. 
Following the spirit of \emph{data-centric AI} \cite{ng2021data}, our methodology can be used with any existing model to find and fix errors in its training set, in order to subsequently train a better model. As better models are invented in the future, all of our proposed methods will more accurately detect label errors in the same data (without modification). 
Our approach can also be easily applied to \emph{multi-task} classification datasets (where the label for some classes may sometimes be missing), by simply ignoring the missing per-class labels when forming our binary one-vs-rest labels $b_i^k$. 

\FloatBarrier

\begin{table}[tb]
\centering
\begin{tabular}{|c|c|}
    \hline
    \textbf{Tag}& \textbf{Accuracy}\\
    \hline
    Eyeglasses& 0.97\\
    \hline
    Wearing\_Earrings& 0.84\\
    \hline
    Wearing\_Hat& 0.97\\
    \hline
    Wearing\_Necklace& 0.87\\
    \hline
    Wearing\_Necktie& 0.93\\
    \hline
    No\_Beard& 0.92\\
    \hline
    Smiling& 0.81\\
    \hline
\end{tabular} 
\caption{Held-out accuracy for each tag (i.e.\ class) obtained by our CelebA multi-label classifier.}
\end{table}

\clearpage

\clearpage
\bibliographystyle{abbrvnat}
\bibliography{multilabel}

\begin{thebibliography}{19}
\providecommand{\natexlab}[1]{#1}
\providecommand{\url}[1]{\texttt{#1}}
\expandafter\ifx\csname urlstyle\endcsname\relax
  \providecommand{\doi}[1]{doi: #1}\else
  \providecommand{\doi}{doi: \begingroup \urlstyle{rm}\Url}\fi

\bibitem[Brodley and Friedl(1999)]{brodley1999identifying}
C.~E. Brodley and M.~A. Friedl.
\newblock Identifying mislabeled training data.
\newblock \emph{Journal of Artificial Intelligence Research}, 11:\penalty0
  131--167, 1999.

\bibitem[Huang et~al.(2019)Huang, Qu, Jia, and Zhao]{huang2019o2u}
J.~Huang, L.~Qu, R.~Jia, and B.~Zhao.
\newblock O2u-net: A simple noisy label detection approach for deep neural
  networks.
\newblock In \emph{Proceedings of the IEEE/CVF International Conference on
  Computer Vision}, 2019.

\bibitem[Kingma and Ba(2015)]{kingma2015adam}
D.~P. Kingma and J.~Ba.
\newblock Adam: A method for stochastic optimization.
\newblock In \emph{International Conference on Learning Representations}, 2015.

\bibitem[Klie et~al.(2022)Klie, Webber, and Gurevych]{klie2022annotation}
J.-C. Klie, B.~Webber, and I.~Gurevych.
\newblock Annotation error detection: Analyzing the past and present for a more
  coherent future.
\newblock \emph{arXiv preprint arXiv:2206.02280}, 2022.

\bibitem[Kuan and Mueller(2022)]{kuan2022}
J.~Kuan and J.~Mueller.
\newblock Model-agnostic label quality scoring to detect real-world label
  errors.
\newblock In \emph{ICML DataPerf Workshop}, 2022.

\bibitem[Lee et~al.(2018)Lee, He, Zhang, and Yang]{lee2017cleannet}
K.-H. Lee, X.~He, L.~Zhang, and L.~Yang.
\newblock Cleannet: Transfer learning for scalable image classifier training
  with label noise.
\newblock In \emph{Proceedings of the IEEE Conference on Computer Vision and
  Pattern Recognition}, 2018.

\bibitem[Liu et~al.(2015)Liu, Luo, Wang, and Tang]{liu2015faceattributes}
Z.~Liu, P.~Luo, X.~Wang, and X.~Tang.
\newblock Deep learning face attributes in the wild.
\newblock In \emph{Proceedings of International Conference on Computer Vision
  (ICCV)}, December 2015.

\bibitem[Müller and Markert(2019)]{muller2019mislabeled}
N.~M. Müller and K.~Markert.
\newblock Identifying mislabeled instances in classification datasets.
\newblock In \emph{International Joint Conference on Neural Networks}, 2019.

\bibitem[Natarajan et~al.(2013)Natarajan, Dhillon, Ravikumar, and
  Tewari]{natarajan2013learning}
N.~Natarajan, I.~S. Dhillon, P.~K. Ravikumar, and A.~Tewari.
\newblock Learning with noisy labels.
\newblock In \emph{Advances in Neural Information Processing Systems}, 2013.

\bibitem[Ng et~al.(2021)Ng, Laird, and He]{ng2021data}
A.~Ng, D.~Laird, and L.~He.
\newblock Data-centric ai competition, 2021.
\newblock \emph{URL https://https-deeplearning-ai. github.
  io/data-centric-comp}, 2021.

\bibitem[Northcutt et~al.(2021{\natexlab{a}})Northcutt, Athalye, and
  Mueller]{northcutt2021labelerrors}
C.~G. Northcutt, A.~Athalye, and J.~Mueller.
\newblock Pervasive label errors in test sets destabilize machine learning
  benchmarks.
\newblock In \emph{Proceedings of the 35th Conference on Neural Information
  Processing Systems Track on Datasets and Benchmarks}, December
  2021{\natexlab{a}}.

\bibitem[Northcutt et~al.(2021{\natexlab{b}})Northcutt, Jiang, and
  Chuang]{northcutt2021confidentlearning}
C.~G. Northcutt, L.~Jiang, and I.~L. Chuang.
\newblock Confident learning: Estimating uncertainty in dataset labels.
\newblock \emph{Journal of Artificial Intelligence Research}, 70:\penalty0
  1373--1411, 2021{\natexlab{b}}.

\bibitem[Pedregosa et~al.(2011)Pedregosa, Varoquaux, Gramfort, Michel, Thirion,
  Grisel, Blondel, Prettenhofer, Weiss, Dubourg, Vanderplas, Passos,
  Cournapeau, Brucher, Perrot, and Duchesnay]{scikit-learn}
F.~Pedregosa, G.~Varoquaux, A.~Gramfort, V.~Michel, B.~Thirion, O.~Grisel,
  M.~Blondel, P.~Prettenhofer, R.~Weiss, V.~Dubourg, J.~Vanderplas, A.~Passos,
  D.~Cournapeau, M.~Brucher, M.~Perrot, and E.~Duchesnay.
\newblock Scikit-learn: Machine learning in {P}ython.
\newblock \emph{Journal of Machine Learning Research}, 12:\penalty0 2825--2830,
  2011.

\bibitem[Reiss et~al.(2020)Reiss, Xu, Cutler, Muthuraman, and
  Eichenberger]{reiss2020identifying}
F.~Reiss, H.~Xu, B.~Cutler, K.~Muthuraman, and Z.~Eichenberger.
\newblock Identifying incorrect labels in the {CoNLL-2003} corpus.
\newblock In \emph{Proceedings of the 24th conference on computational natural
  language learning}, pages 215--226, 2020.

\bibitem[Rottmann and Reese(2022)]{segmentation}
M.~Rottmann and M.~Reese.
\newblock Automated detection of label errors in semantic segmentation datasets
  via deep learning and uncertainty quantification.
\newblock \emph{arXiv preprint arXiv:2207.06104}, 2022.

\bibitem[Song et~al.(2022)Song, Kim, Park, Shin, and Lee]{song2022survey}
H.~Song, M.~Kim, D.~Park, Y.~Shin, and J.-G. Lee.
\newblock Learning from noisy labels with deep neural networks: A survey.
\newblock \emph{IEEE Transactions on Neural Networks and Learning Systems},
  2022.

\bibitem[Tan and Le(2019)]{tan2019efficientnet}
M.~Tan and Q.~Le.
\newblock Efficientnet: Rethinking model scaling for convolutional neural
  networks.
\newblock In \emph{International Conference on Machine Learning}, 2019.

\bibitem[Wang and Mueller(2022)]{token}
W.-C. Wang and J.~Mueller.
\newblock Detecting label errors in token classification data.
\newblock \emph{arXiv preprint arXiv:2210.03920}, 2022.

\bibitem[Wightman(2019)]{rw2019timm}
R.~Wightman.
\newblock Pytorch image models.
\newblock \url{https://github.com/rwightman/pytorch-image-models}, 2019.

\end{thebibliography}

\clearpage \newpage
\beginsupplement
\onecolumn
\appendix

\def\toptitlebar{\hrule height1pt \vskip .2in} 
\def\bottomtitlebar{\vskip .22in \hrule height1pt \vskip .3in} 

\setlength{\footskip}{20pt}  
\begin{center}
{\LARGE \bf Appendix }
\end{center}
\FloatBarrier

\section{Details for Label Quality Score Benchmark}
\label{sec:datadetails}

Here we describe our \emph{large}/\emph{small} groups of multi-label classification datasets used for evaluation. The ten datasets in each group were each generated from the same underlying distribution with different random seeds. We produced each dataset via the  \texttt{make\_multilabel\_classification} data generator from the \texttt{sklearn} package\footnote{\url{https://scikit-learn.org/stable/modules/generated/sklearn.datasets.make_multilabel_classification.html}} \citep{scikit-learn}. This method produces bag of words features (our datasets had an expected word count of 500). 
Table \ref{tab: dataset groups} lists various differences between our \emph{Large} and \emph{Small} datasets.

We introduce class-wise label noise in the given labels by randomly generating class noise matrices with traces $T_k$ for class $k$, described by \cref{eq:gammatrace,eq:tracemax}:
\begin{equation}
    Y_k = (1 - X_k) \left(1 - \frac{\exp\left( \frac{-(k_\mathrm{argsort}-1)^2}{K} \right)}{2K} \right)
    \label{eq:gammatrace}
\end{equation}
\begin{equation}
    T_k = \max \{2Y_k, 2 - 2Y_k \}
    \label{eq:tracemax}
\end{equation}
where $X_k \sim \Gamma(\kappa, \theta)$ is drawn from a gamma distribution  with $(\kappa, \theta) = (2, 0.01)$ and $k_\mathrm{argsort}$ is the index of $X_k$ in ${X = (X_1, \dots, X_K)}$ after being sorted in ascending order. The expression in the latter parentheses in \cref{eq:gammatrace} reweights the samples exponentially based on their relative ordering to emphasize classes with the worst label noise. Each example $i$ is limited to at most 3 errors in the individual per class annotations $b_i^1, ..., b_i^K$.
\begin{figure}[h]
\centering
\includegraphics[width=0.75\textwidth]{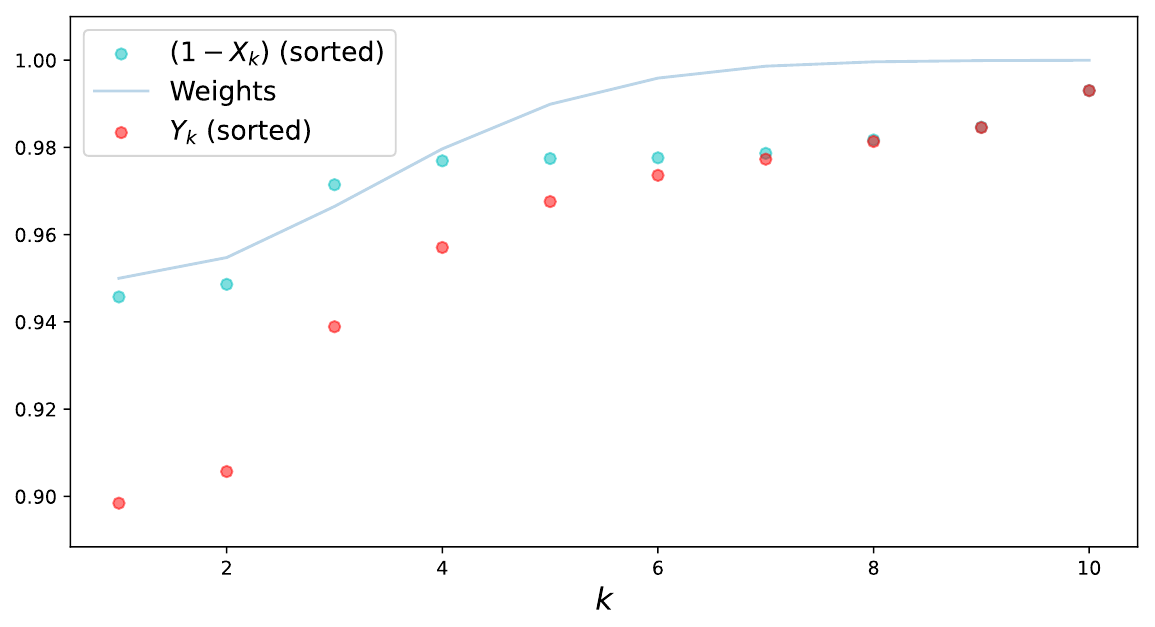}
\vspace*{-0.3cm}
\caption{Average traces of noise matrices for a dataset with $K=10$ classes, sampled via  \cref{eq:gammatrace}.}
\end{figure}
\begin{table}[h]
\centering
\caption{Summaries of the two groups of synthetically generated datasets}
\vspace*{0.2cm}
\begin{tabular}{lll}
\textbf{}                                                           & Small                     & \multicolumn{1}{c}{Large}  \\ \hline
\multicolumn{1}{|l|}{\textbf{Number of samples}}                    & \multicolumn{1}{l|}{5000} & \multicolumn{1}{l|}{30000} \\ \hline
\multicolumn{1}{|l|}{\textbf{Number of test samples}}                    & \multicolumn{1}{l|}{1000} & \multicolumn{1}{l|}{7500} \\ \hline
\multicolumn{1}{|l|}{\textbf{Number of features}}                   & \multicolumn{1}{l|}{3}    & \multicolumn{1}{l|}{20}    \\ \hline
\multicolumn{1}{|l|}{\textbf{Number of classes}}                    & \multicolumn{1}{l|}{4}    & \multicolumn{1}{l|}{50}    \\ \hline
\multicolumn{1}{|l|}{\textbf{Expected number of class assignments}} & \multicolumn{1}{l|}{2}    & \multicolumn{1}{l|}{5}     \\ \hline
\end{tabular}
\label{tab: dataset groups}
\end{table}

\clearpage
\section{Additional Results for Label Quality Score Benchmark}
\label{sec:additionalresults}

\begin{table}[h]
    \centering
    \caption{The average-per-class accuracies and Jaccard scores of different models trained and tested with either true or noisy labels for each group of datasets. Values reported are the mean over the 10 datasets in each group. In parenthesis is the corresponding standard deviation to the precision of the least significant digit of the mean value.}
\begin{tabular}{llllcc}
                       &                                      &                        &                            & \textbf{Average accuracy}     & \textbf{Jaccard score}       \\
\textbf{Datasets}      & \textbf{Classifier}                  & \textbf{Train labels}  & \textbf{Test labels}       &                               &                              \\ \hline
\multirow{8}{*}{Small} & \multirow{4}{*}{Logistic regression} & \multirow{2}{*}{Noisy} & \multicolumn{1}{l|}{Noisy} & \multicolumn{1}{c|}{0.76(3)}  & \multicolumn{1}{c|}{0.62(4)} \\ \cline{5-6} 
                       &                                      &                        & \multicolumn{1}{l|}{True}  & \multicolumn{1}{c|}{0.81(4)}  & \multicolumn{1}{c|}{0.69(6)} \\ \cline{3-6} 
                       &                                      & \multirow{2}{*}{True}  & \multicolumn{1}{l|}{Noisy} & \multicolumn{1}{c|}{0.75(3)}  & \multicolumn{1}{c|}{0.61(5)} \\ \cline{5-6} 
                       &                                      &                        & \multicolumn{1}{l|}{True}  & \multicolumn{1}{c|}{0.81(3)}  & \multicolumn{1}{c|}{0.69(6)} \\ \cline{2-6} 
                       & \multirow{4}{*}{Random forest}       & \multirow{2}{*}{Noisy} & \multicolumn{1}{l|}{Noisy} & \multicolumn{1}{c|}{0.81(3)}  & \multicolumn{1}{c|}{0.69(5)} \\ \cline{5-6} 
                       &                                      &                        & \multicolumn{1}{l|}{True}  & \multicolumn{1}{c|}{0.88(4)}  & \multicolumn{1}{c|}{0.81(6)} \\ \cline{3-6} 
                       &                                      & \multirow{2}{*}{True}  & \multicolumn{1}{l|}{Noisy} & \multicolumn{1}{c|}{0.82(3)}  & \multicolumn{1}{c|}{0.72(5)} \\ \cline{5-6} 
                       &                                      &                        & \multicolumn{1}{l|}{True}  & \multicolumn{1}{c|}{0.90(4)}  & \multicolumn{1}{c|}{0.86(6)} \\ \hline
\multirow{8}{*}{Large} & \multirow{4}{*}{Logistic regression} & \multirow{2}{*}{Noisy} & \multicolumn{1}{l|}{Noisy} & \multicolumn{1}{c|}{0.906(3)} & \multicolumn{1}{c|}{0.30(2)} \\ \cline{5-6} 
                       &                                      &                        & \multicolumn{1}{l|}{True}  & \multicolumn{1}{c|}{0.922(2)} & \multicolumn{1}{c|}{0.36(2)} \\ \cline{3-6} 
                       &                                      & \multirow{2}{*}{True}  & \multicolumn{1}{l|}{Noisy} & \multicolumn{1}{c|}{0.907(3)} & \multicolumn{1}{c|}{0.31(2)} \\ \cline{5-6} 
                       &                                      &                        & \multicolumn{1}{l|}{True}  & \multicolumn{1}{c|}{0.923(2)} & \multicolumn{1}{c|}{0.38(2)} \\ \cline{2-6} 
                       & \multirow{4}{*}{Random forest}       & \multirow{2}{*}{Noisy} & \multicolumn{1}{l|}{Noisy} & \multicolumn{1}{c|}{0.903(3)} & \multicolumn{1}{c|}{0.25(2)} \\ \cline{5-6} 
                       &                                      &                        & \multicolumn{1}{l|}{True}  & \multicolumn{1}{c|}{0.919(2)} & \multicolumn{1}{c|}{0.31(2)} \\ \cline{3-6} 
                       &                                      & \multirow{2}{*}{True}  & \multicolumn{1}{l|}{Noisy} & \multicolumn{1}{c|}{0.903(3)} & \multicolumn{1}{c|}{0.25(2)} \\ \cline{5-6} 
                       &                                      &                        & \multicolumn{1}{l|}{True}  & \multicolumn{1}{c|}{0.919(2)} & \multicolumn{1}{c|}{0.31(2)} \\ \cline{4-6} 
\end{tabular}
    \label{tab: model accuracies}
\end{table}

\begin{figure}[!h] 
\begin{center}
\textbf{Logistic Regression} \\[0.2cm]
\includegraphics[width=0.85\textwidth, trim={5cm 5cm 5cm 5cm},clip]{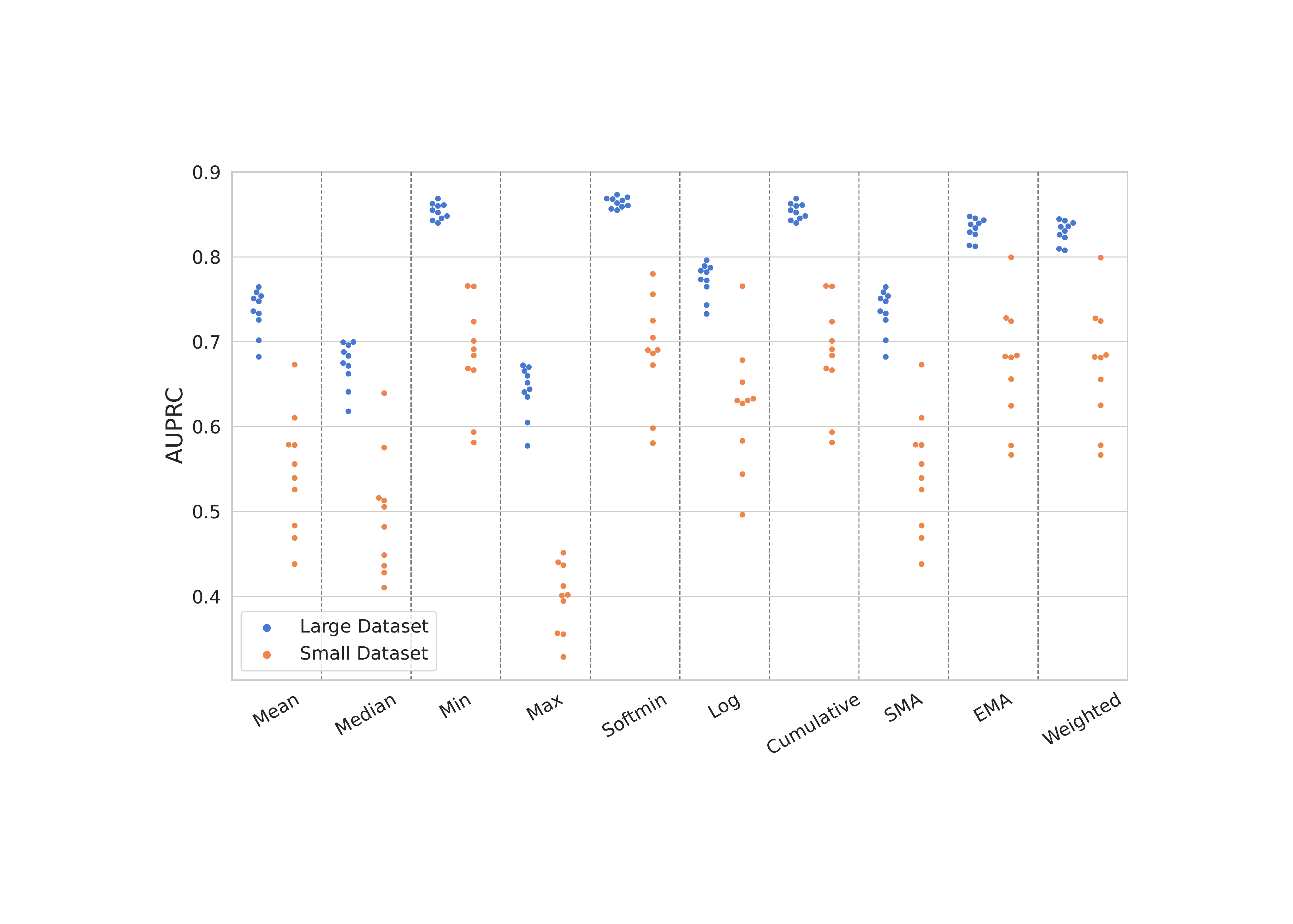}
\\[0.5cm]
\textbf{Random Forest} \\[0.2cm]
\includegraphics[width=0.85\textwidth, trim={5cm 5cm 5cm 5cm},clip]{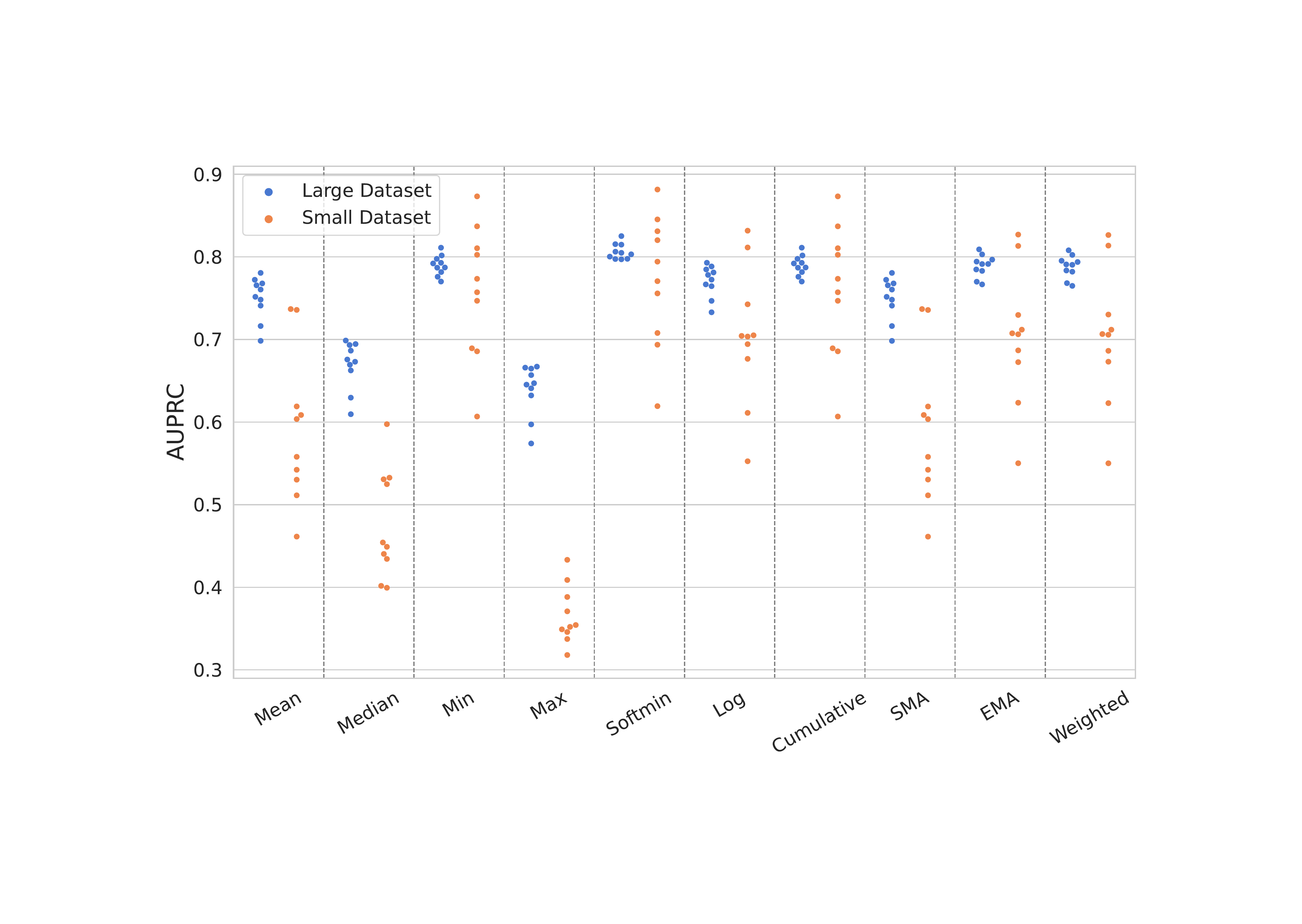}
\end{center}
\vspace*{-0.3cm}
\caption{AUPRC for detecting mislabeled examples  achieved by various overall label quality scores $\scoreEg{i}$ for each dataset (shown as dot).
We show results based on predicted class probabilities from both a Logistic Regression model and a Random Forest model.}
\label{fig:auprc}
\end{figure}

\begin{figure}[!h] 
\begin{center}
\textbf{Logistic Regression} \\[0.2cm]
\includegraphics[width=0.85\textwidth, trim={5cm 5cm 5cm 5cm},clip]{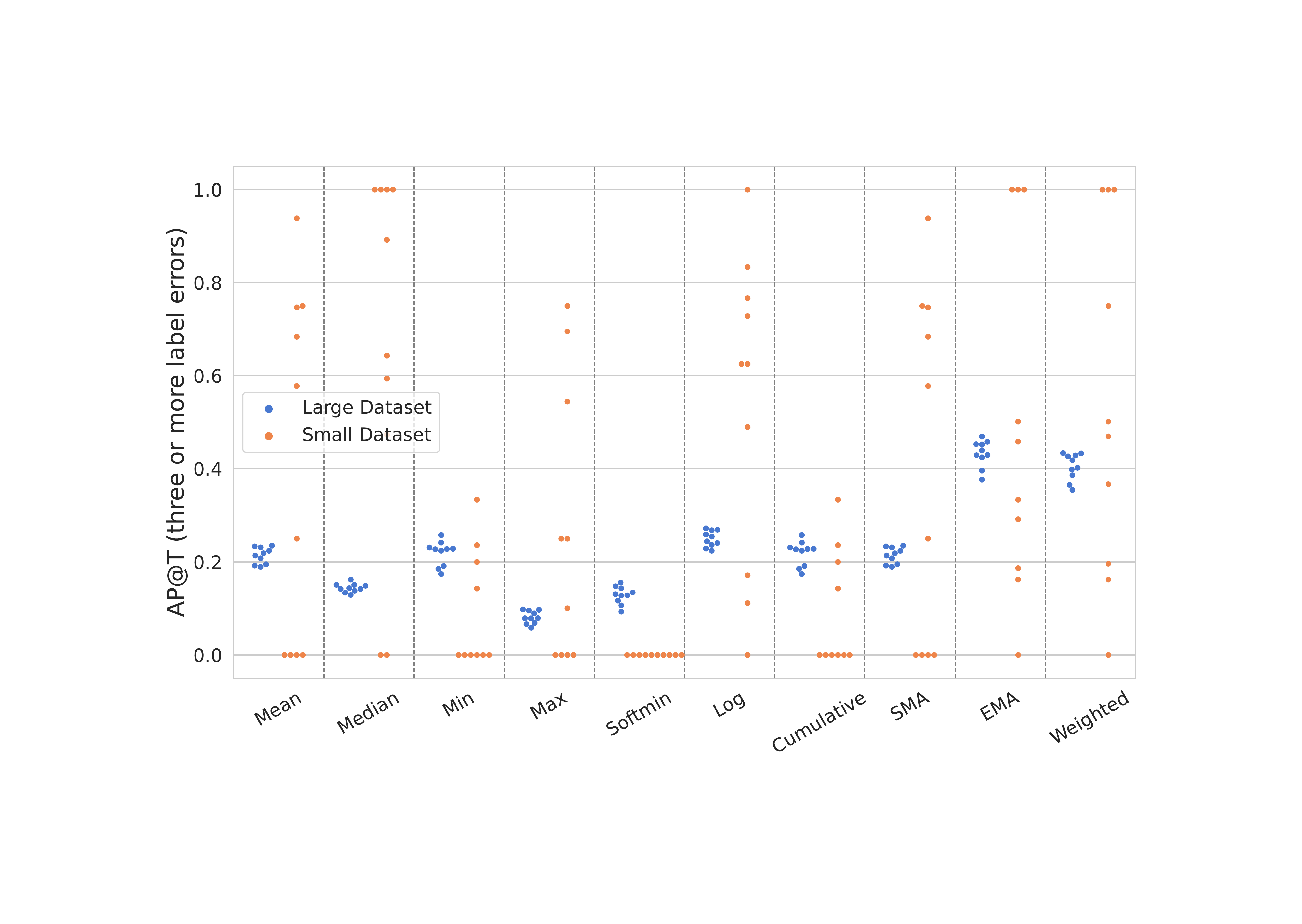}
\\[0.5cm]
\textbf{Random Forest} \\[0.2cm]
\includegraphics[width=0.85\textwidth, trim={5cm 5cm 5cm 5cm},clip]{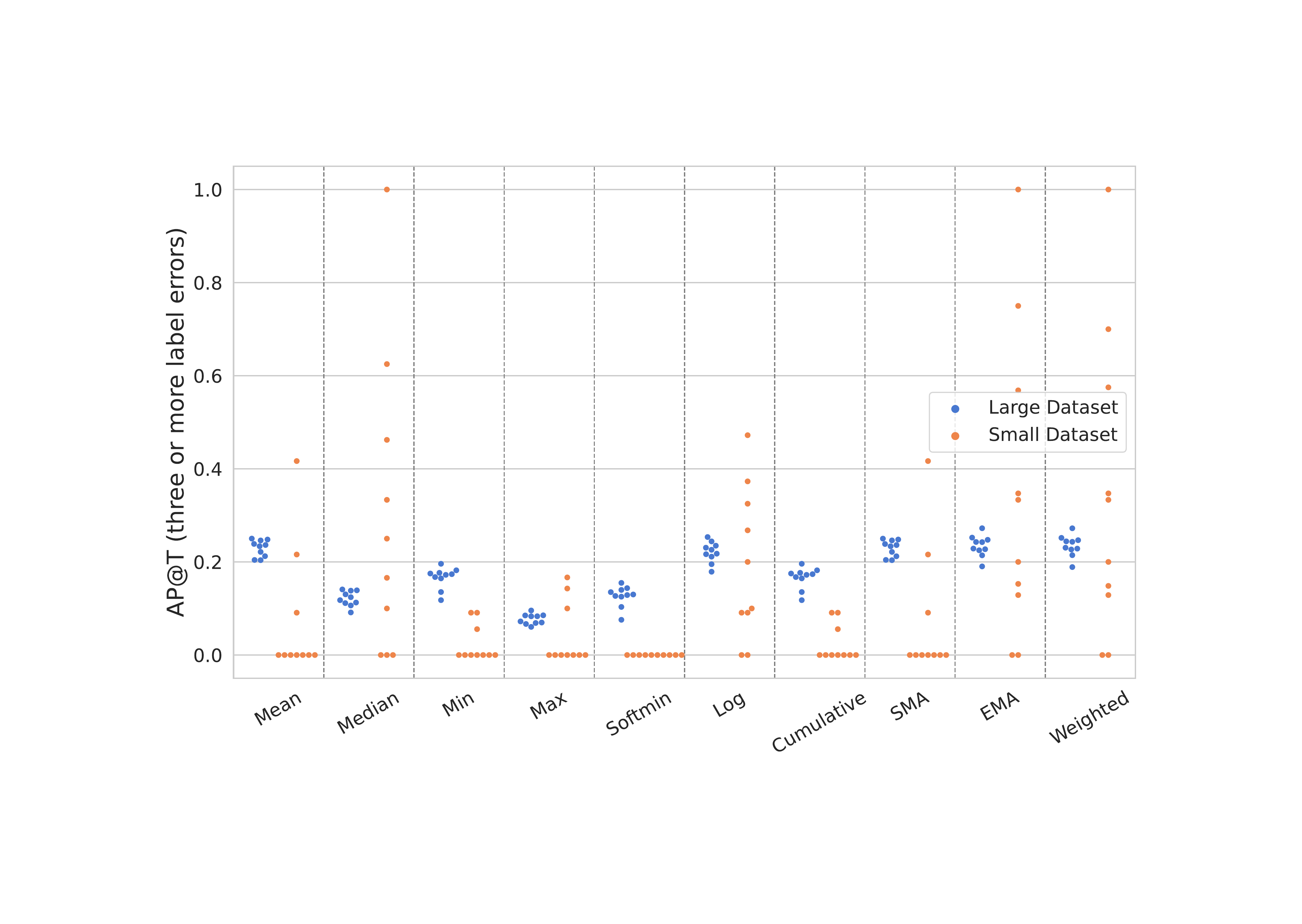}
\end{center}
\vspace*{-0.3cm}
\caption{Average 3-Precision @ $T$ achieved by various overall label quality scores $\scoreEg{i}$, where $T$ is the number of mislabeled examples in each dataset.
We show results based on predicted class probabilities from both a Logistic Regression model and a Random Forest model.}
\label{fig:ap3_numerr}
\end{figure}

\begin{figure}[!h] 
\begin{center}
\textbf{Logistic Regression} \\[0.2cm]
\includegraphics[width=0.85\textwidth, trim={5cm 5cm 5cm 5cm},clip]{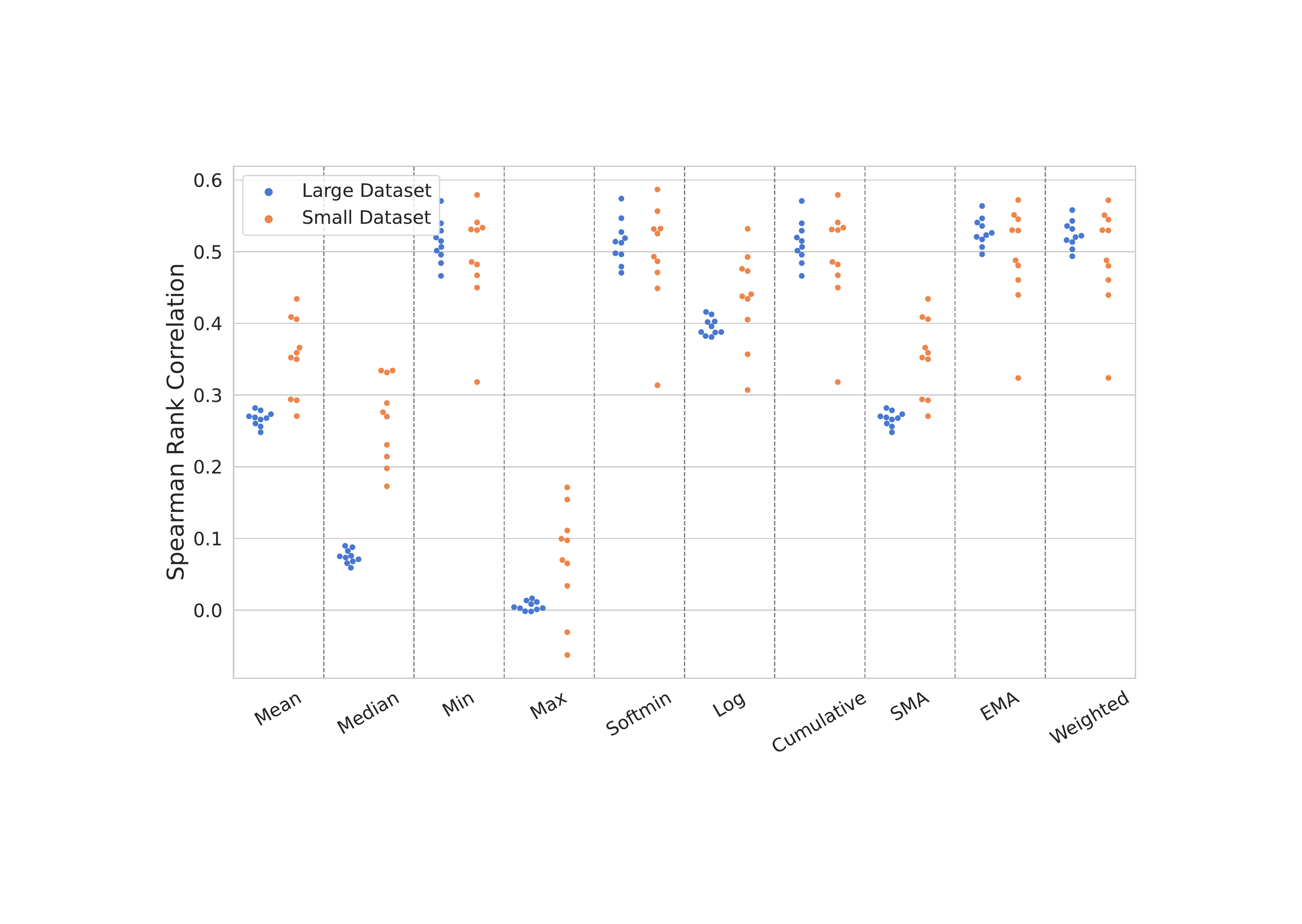}
\\[0.5cm]
\textbf{Random Forest} \\[0.2cm]
\includegraphics[width=0.85\textwidth, trim={5cm 5cm 5cm 5cm},clip]{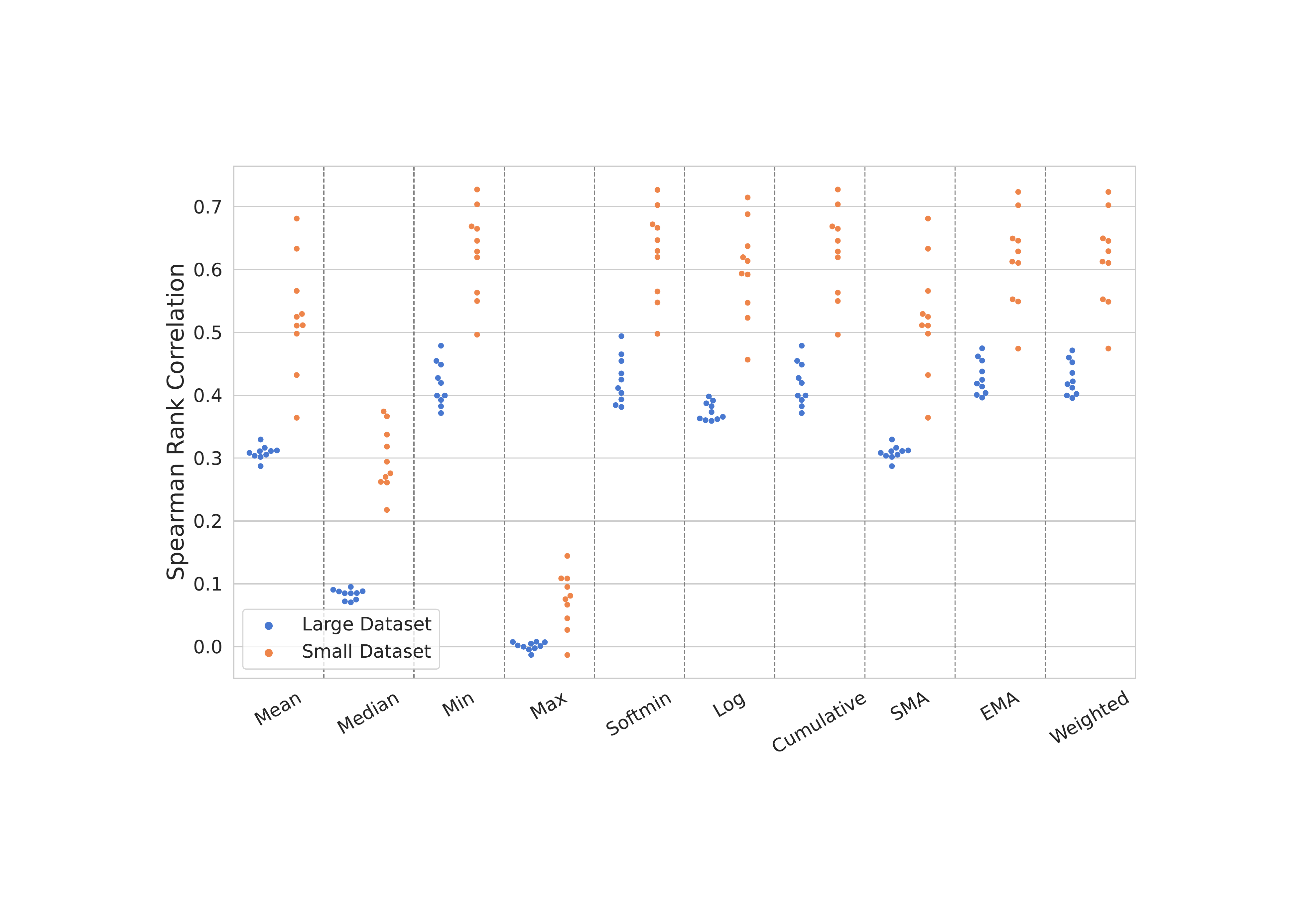}
\end{center}
\vspace*{-0.3cm}
\caption{Spearman correlation between label quality scores $\scoreEg{i}$ (produced via various methods) and number of class annotations $b_i^1, ..., b_i^K$ which are incorrect per example $i$.
We show results based on predicted class probabilities from both a Logistic Regression model and a Random Forest model.}
\label{fig:spearman}
\end{figure}

\begin{figure}[!h] 
\begin{center}
\textbf{Logistic Regression} \\[0.2cm]
\includegraphics[width=0.9\textwidth, trim={5cm 4cm 5cm 5cm},clip]{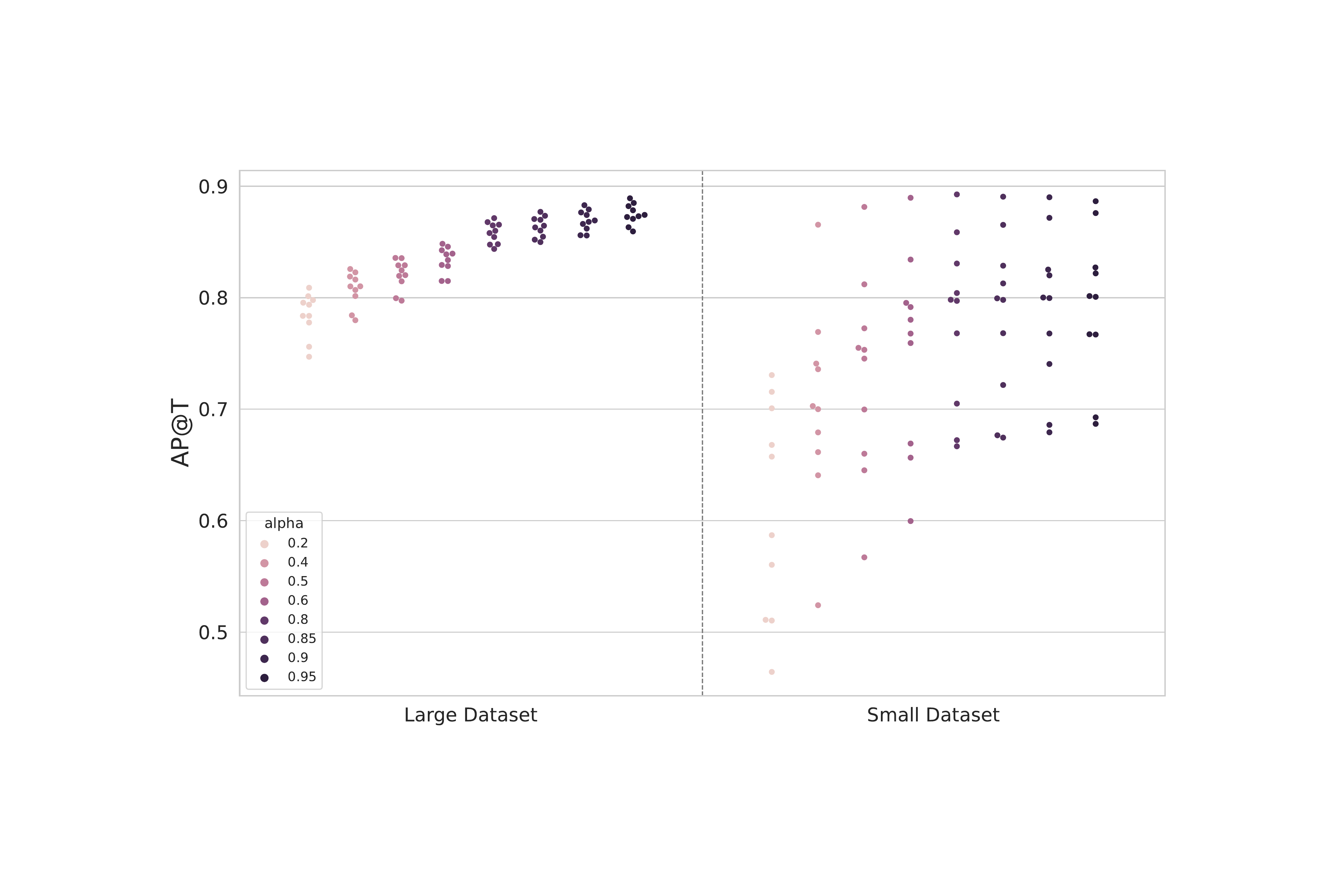}
\textbf{Random Forest} \\[0.2cm]
\includegraphics[width=0.9\textwidth, trim={5cm 4cm 5cm 5cm},clip]{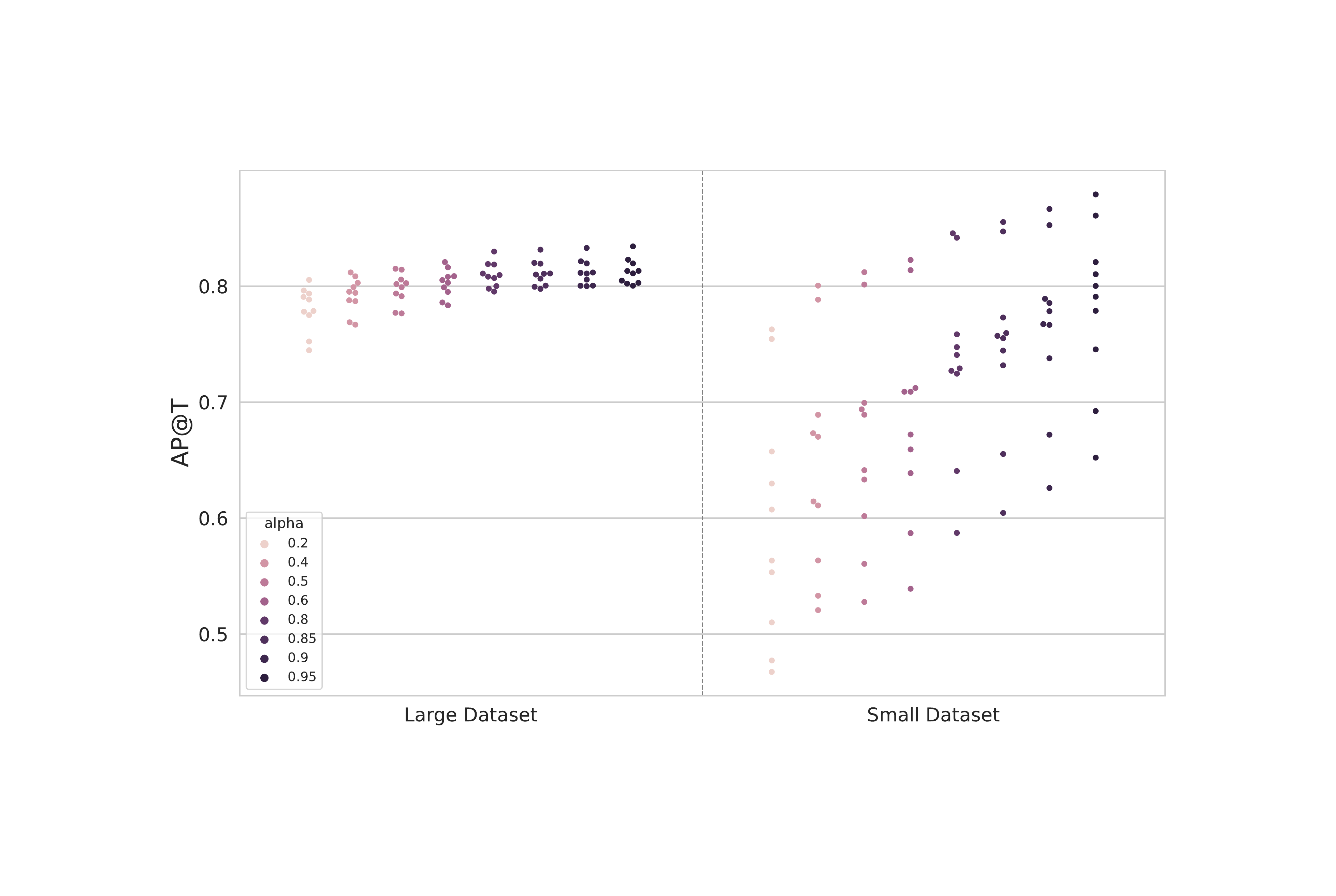}
\end{center}
\vspace*{-0.7cm}
\caption{Average Precision @ $T$ achieved by our EMA label quality scoring method with different values of $\alpha$. Here $T$ is the number of mislabeled examples in each dataset, and all examples with any sort of error in their label are counted as positive hits. 
We show results based on  predicted class probabilities from both a Logistic Regression model and a Random Forest model.}
\label{fig:alpha1}
\end{figure}

\begin{figure}[!h] 
\begin{center}
\textbf{Logistic Regression} \\[0.2cm]
\includegraphics[width=0.9\textwidth, trim={5cm 4cm 5cm 5cm},clip]{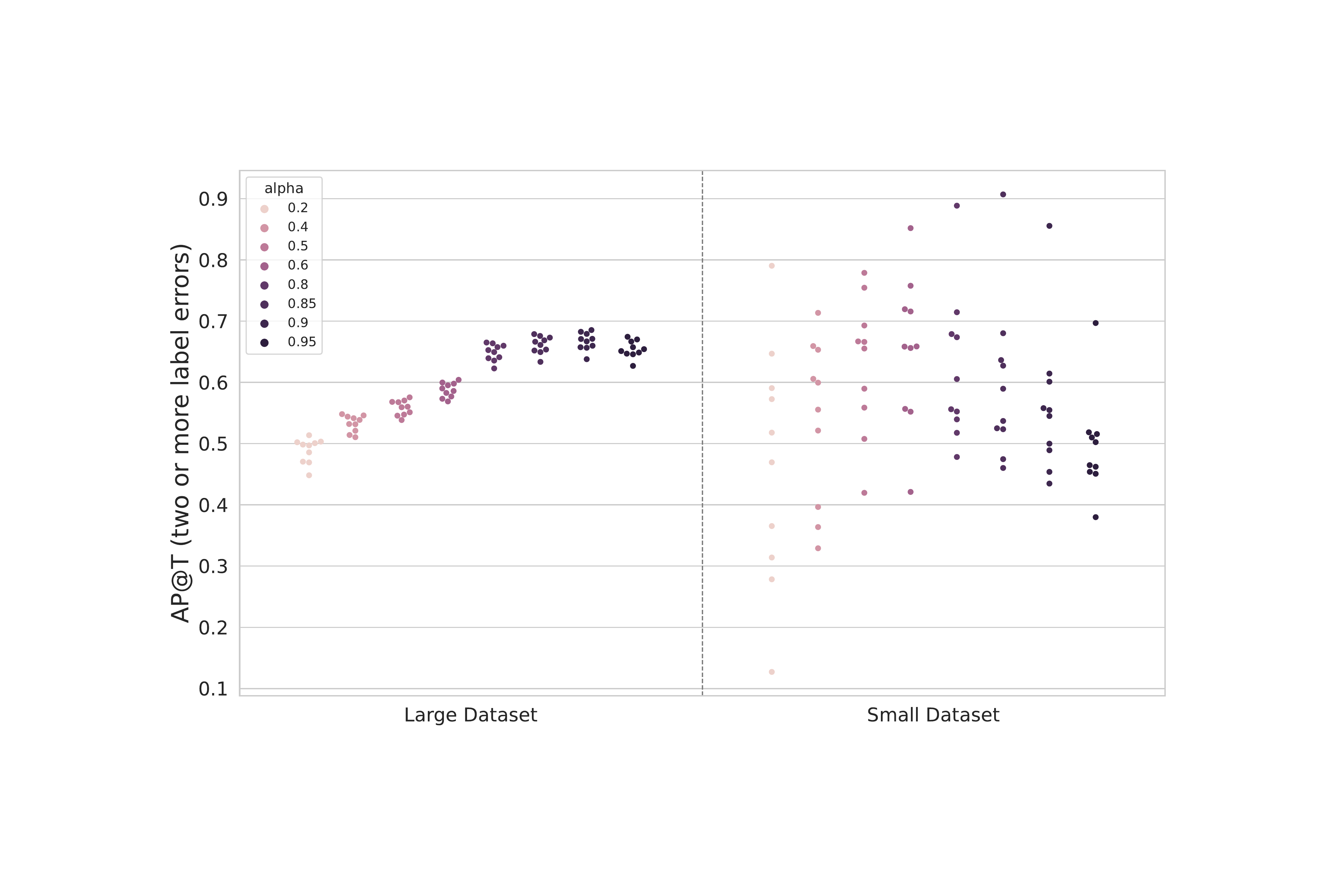}
\\
\textbf{Random Forest} \\[0.2cm]
\includegraphics[width=0.9\textwidth, trim={5cm 4cm 5cm 5cm},clip]{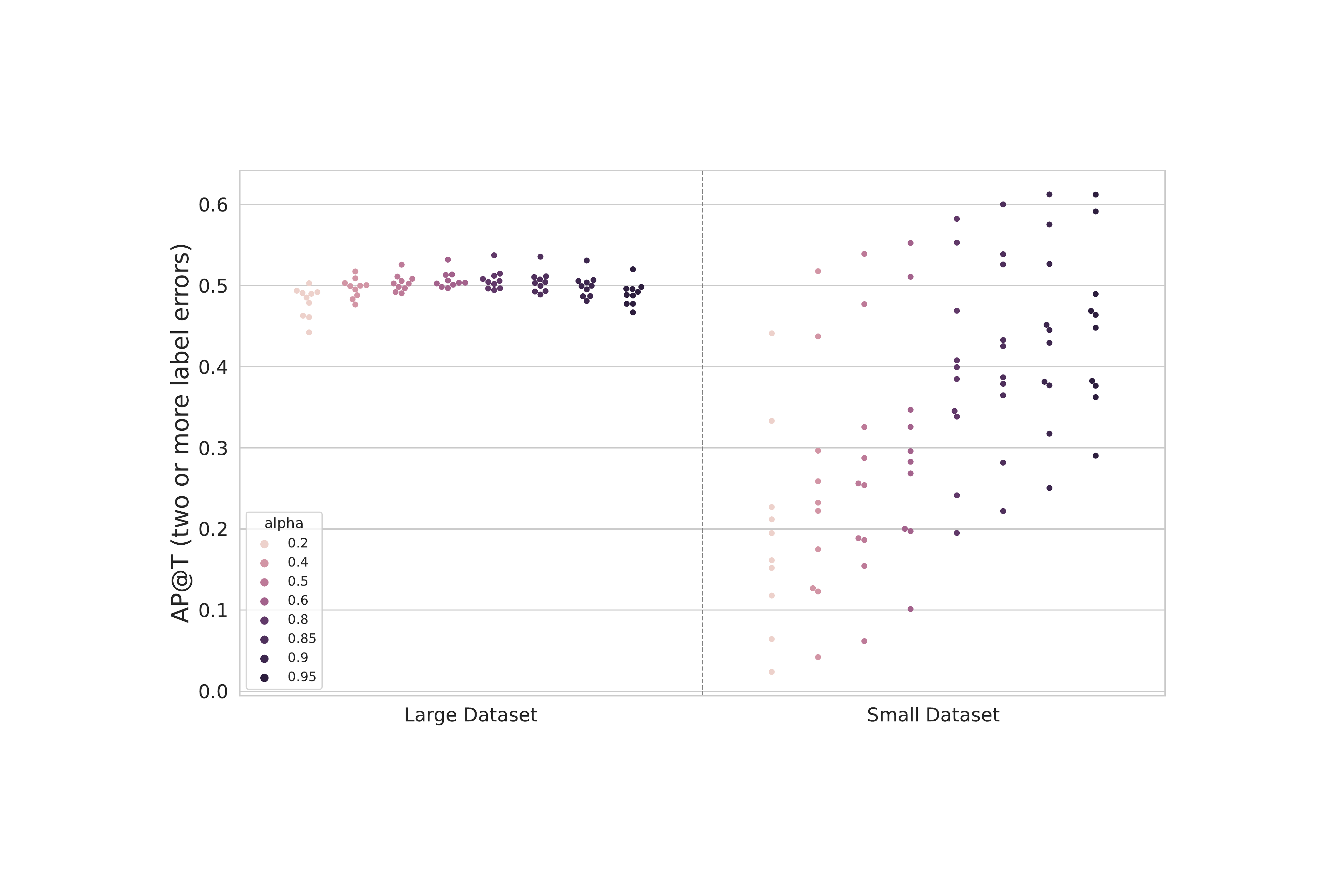}
\end{center}
\vspace*{-0.7cm}
\caption{Average 2-Precision @ $T$ achieved by our EMA label quality scoring method with different values of $\alpha$. In 2-Precision, an example is only counted as a positive hit if its label contains at least 2 misannotated classes. $T$ is the number of mislabeled examples in each dataset. We show results based on  predicted class probabilities from both a Logistic Regression model and a Random Forest model.}
\label{fig:alpha2}
\end{figure}

\begin{figure}[!h] 
\begin{center}
\textbf{Logistic Regression} \\[0.2cm]
\includegraphics[width=0.9\textwidth, trim={5cm 4cm 5cm 5cm},clip]{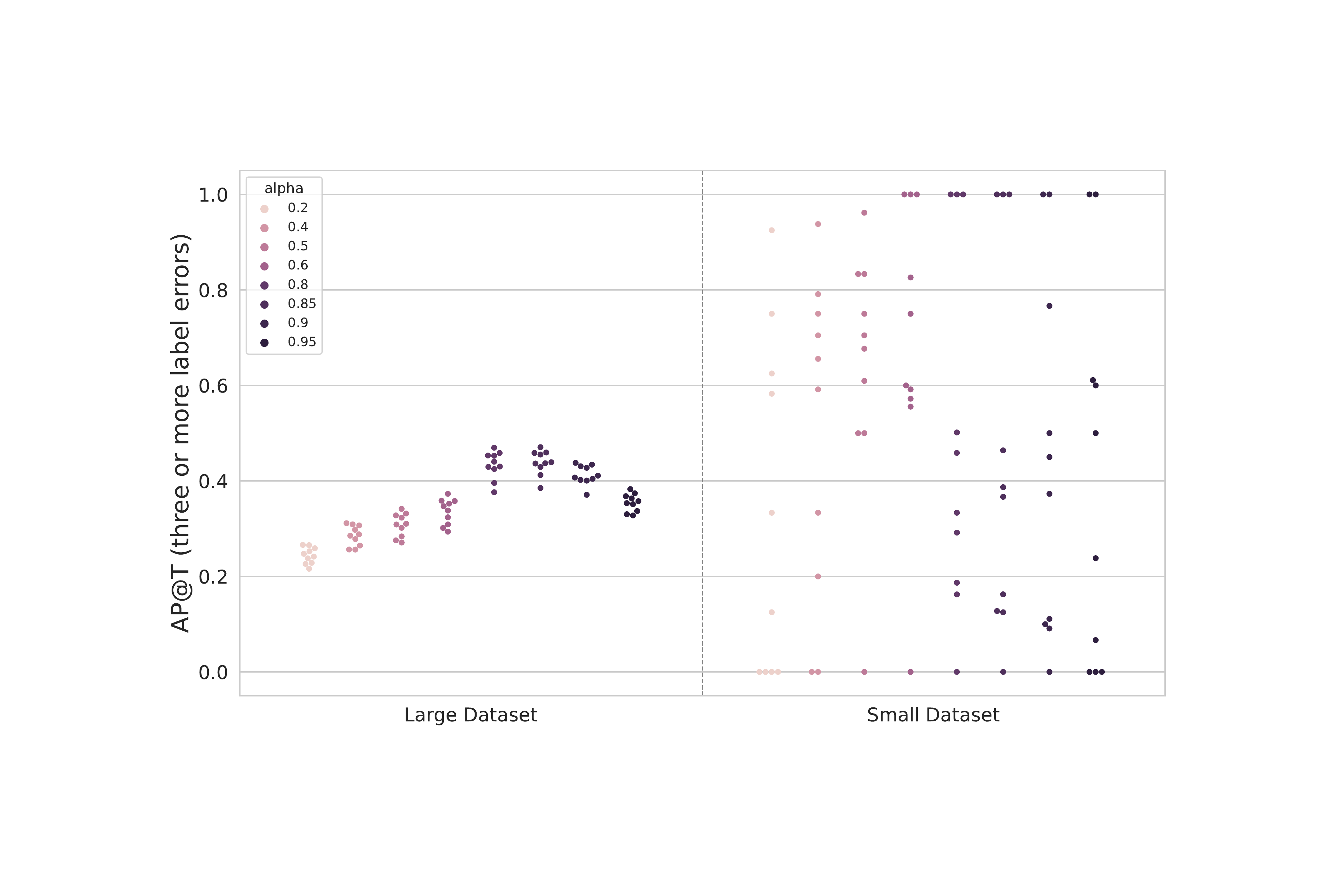}
\\
\textbf{Random Forest} \\[0.2cm]
\includegraphics[width=0.9\textwidth, trim={5cm 4cm 5cm 5cm},clip]{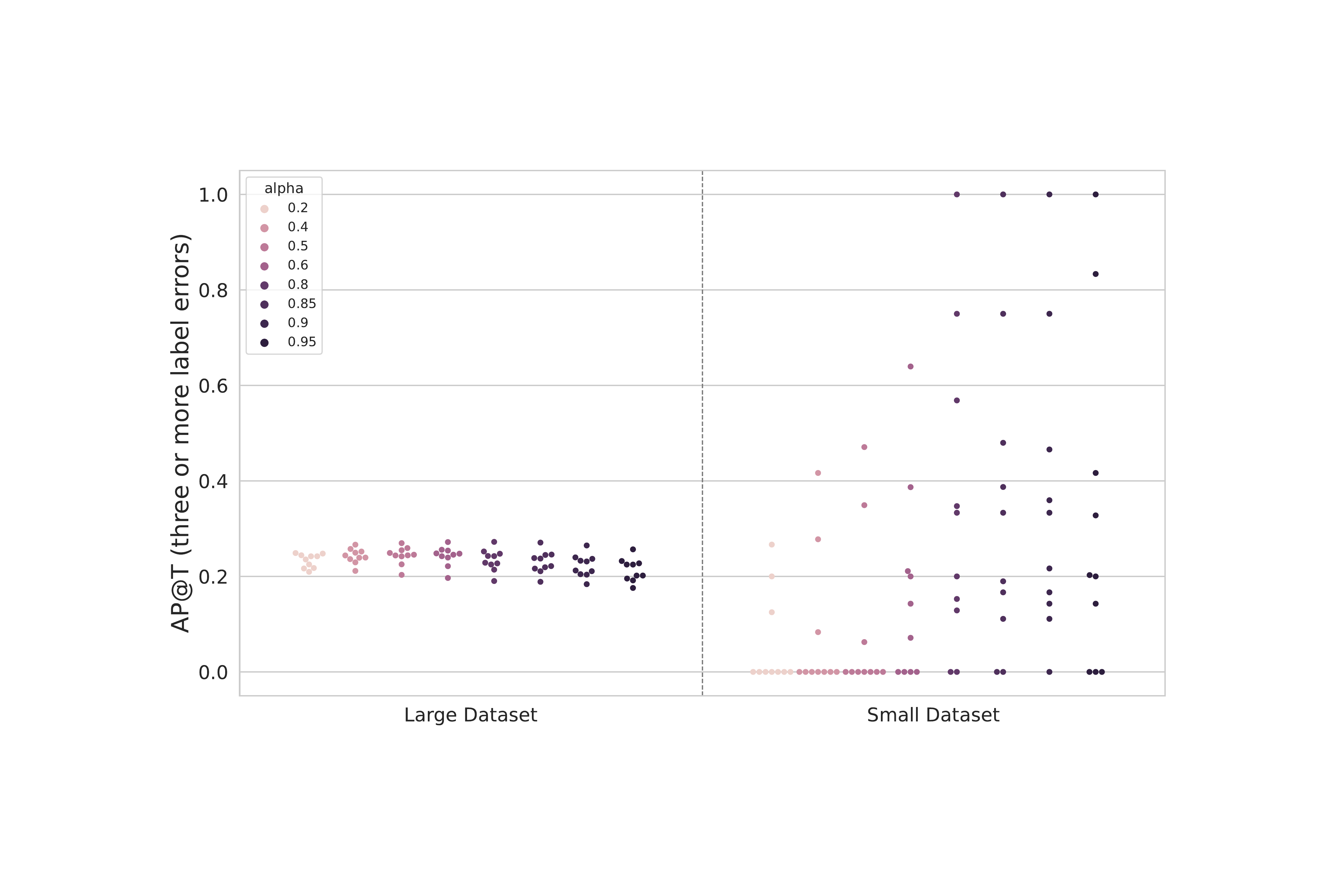}
\end{center}
\vspace*{-0.7cm}
\caption{Average 3-Precision @ $T$ achieved by our EMA label quality scoring method with different values of $\alpha$. In 3-Precision, an example is only counted as a positive hit if its label contains at least 3 misannotated classes. $T$ is the number of mislabeled examples in each dataset. We show results based on  predicted class probabilities from both a Logistic Regression model and a Random Forest model.}
\label{fig:alpha3}
\end{figure}
\begin{figure}[!h] 
\begin{center}
\textbf{Logistic Regression} \\[0.2cm]
\includegraphics[width=0.9\textwidth, trim={5cm 4cm 5cm 5cm},clip]{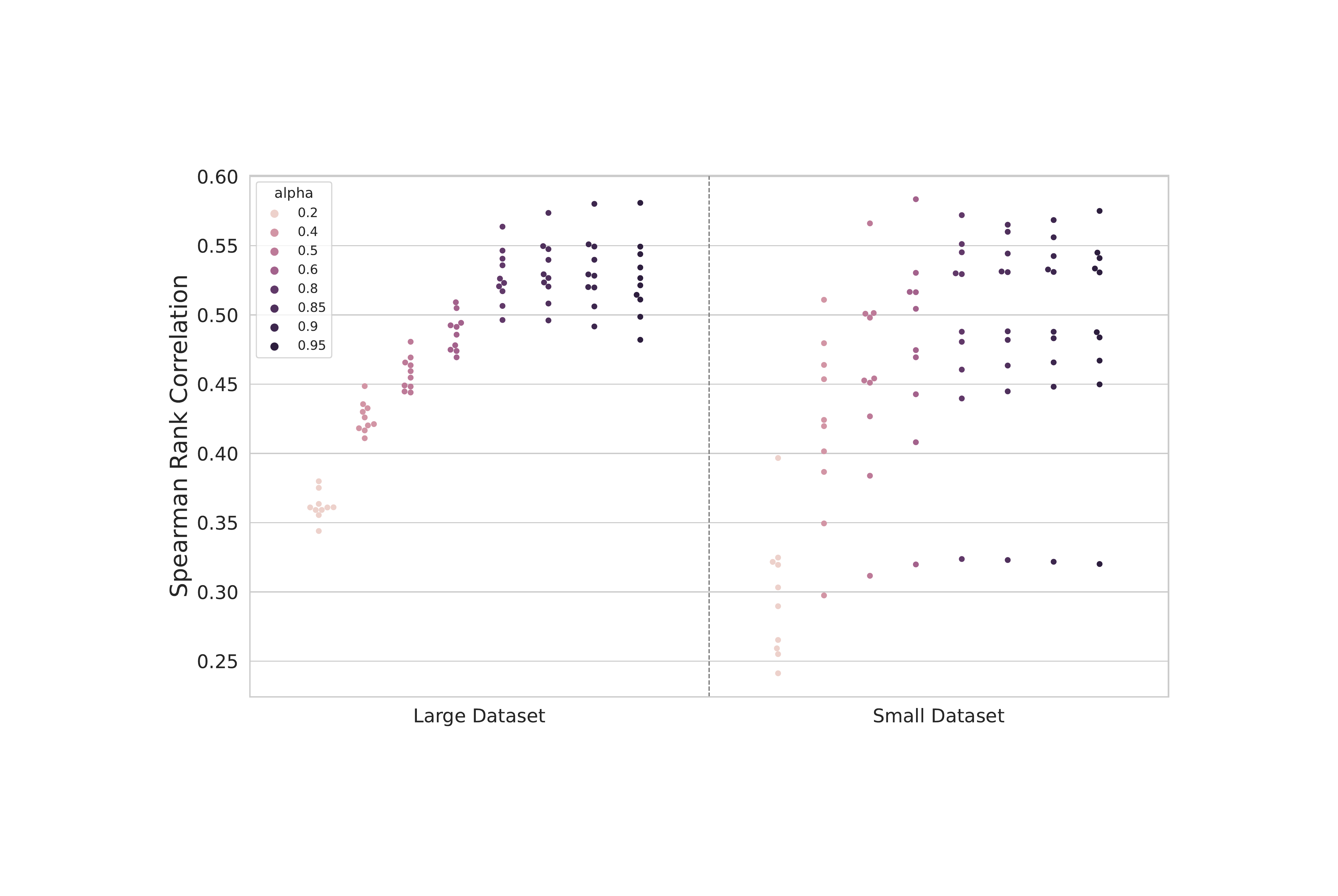}
\\
\textbf{Random Forest} \\[0.2cm]
\includegraphics[width=0.9\textwidth, trim={5cm 4cm 5cm 5cm},clip]{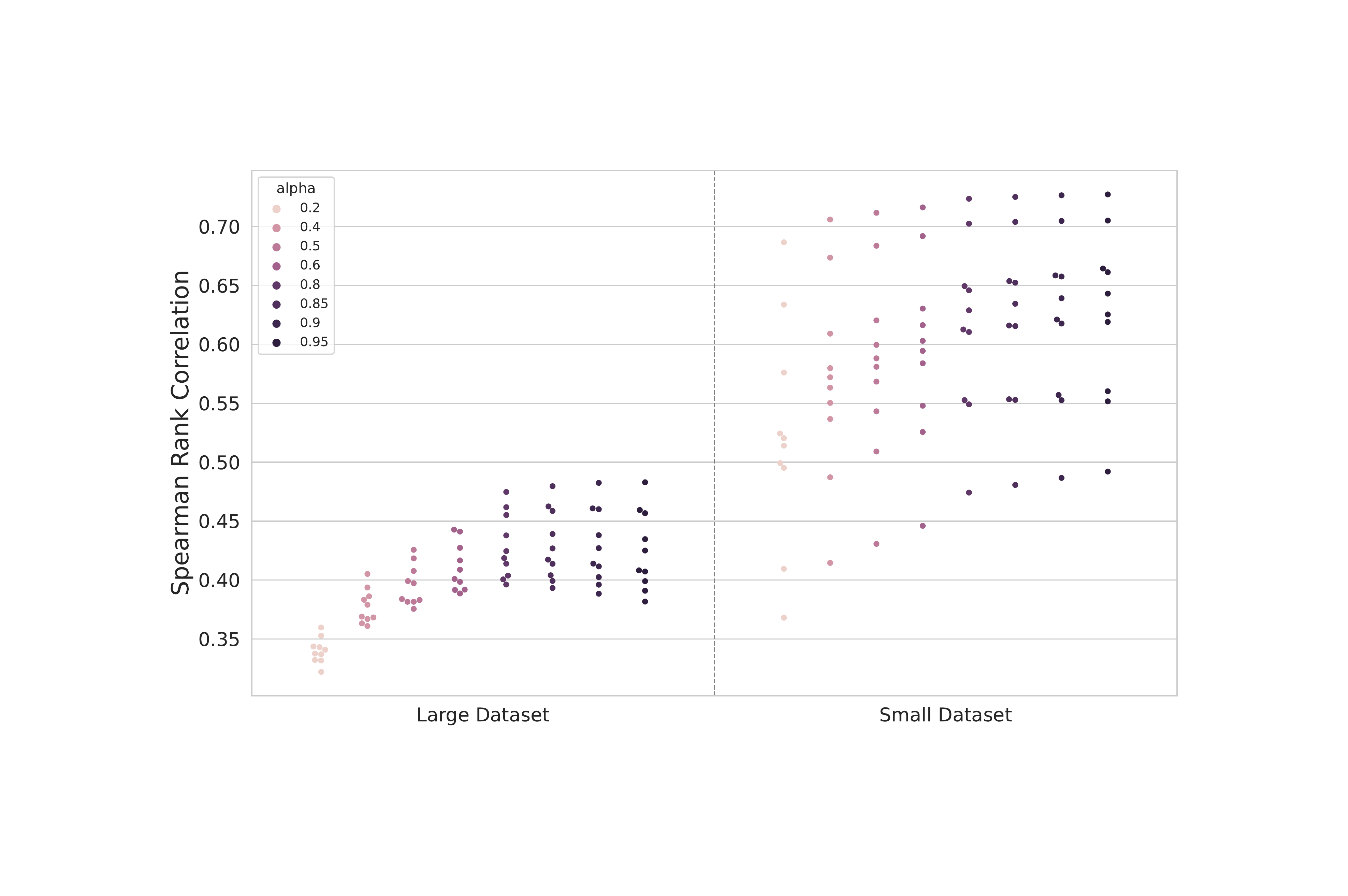}
\end{center}
\vspace*{-0.7cm}
\caption{Spearman correlation between label quality scores $\scoreEg{i}$ and number of class annotations $b_i^1, ..., b_i^K$ which are incorrect per example $i$. The scores $\scoreEg{i}$ are produced via our EMA method with different values of $\alpha$. We show results based on predicted class probabilities from both a Logistic Regression model and a Random Forest model.}
\label{fig:alpha4}
\end{figure}

\end{document}